# A Different Approach to AI Safety

## Proceedings from the Columbia Convening on AI Openness and Safety

Camille François*[1], Ludovic Péran*[1], Ayah Bdeir*[2], Nouha Dziri[3], Will Hawkins[4], Yacine Jernite[5], Sayash Kapoor[6], Juliet Shen[1,15], Heidy Khlaaf[7], Kevin Klyman[8], Nik Marda[2], Marie Pellat[9], Deb Raji[2], Divya Siddarth[10], Aviya Skowron[11], Joseph Spisak[12], Madhulika Srikumar[13], Victor Storchan[2], Audrey Tang[14], Jen Weedon[1]

[1]Columbia University, [2]Mozilla, [3]Ai2, [4]Google DeepMind, [5]HuggingFace, [6]Princeton CITP, [7]AI Now Institute, [8]Stanford CRFM, [9]Mistral, [10]Collective Intelligence Project, [11]EleutherAI, [12]Meta, [13]Partnership on AI, [14]Taiwan Digital Affairs, [15]ROOST

Correspondence to Camille François <cmf2157@columbia.edu>, Ludovic Péran <ludovic.peran@gmail.com>, Ayah Bdeir <ayahbdeir@gmail.com>

# Abstract

The rapid rise of open-weight and open-source foundation models is intensifying the obligation and reshaping the opportunity to make AI systems safe. This paper reports outcomes from the Columbia Convening on AI Openness and Safety (San Francisco, 19 Nov 2024) and its six-week preparatory programme involving more than forty-five researchers, engineers, and policy leaders from academia, industry, civil society, and government. Using a participatory, solutions-oriented process, the working groups produced (i) a research agenda at the intersection of safety and open source AI; (ii) a mapping of existing and needed technical interventions and open source tools to safely and responsibly deploy open foundation models across the AI development workflow and (iii) a mapping of the content safety filter ecosystem with a proposed roadmap for future research and development. We find that openness—understood as transparent weights, interoperable tooling, and public governance—can enhance safety by enabling independent scrutiny, decentralised mitigation, and culturally plural oversight. However, significant gaps persist: scarce multimodal and multilingual benchmarks, limited defences against prompt-injection and compositional attacks in agentic systems, and insufficient participatory mechanisms for communities most affected by AI harms. The paper concludes with a roadmap of five priority research directions, emphasising participatory inputs, future-proof content filters, ecosystem-wide safety infrastructure, rigorous agentic safeguards, and expanded harm taxonomies. These recommendations informed the February 2025 French AI Action Summit and lay groundwork for an open, plural, and accountable AI safety discipline.

## Executive Summary

On November 19, 2024, Mozilla and Columbia University's Institute of Global Politics hosted the Columbia Convening on AI Openness and Safety in San Francisco. This event was part of the ongoing Columbia Convening series on AI and Openness, which launched in October 2023 alongside the UK Safety Summit with an [open letter](#) coordinated by Mozilla and Columbia signed by more than 1,800 leading experts and community members declaring that, "when it comes to AI Safety and Security, openness is an antidote not a poison." Shortly after, both organizations have committed to facilitate an ongoing, dynamic and inclusive dialogue about what "open" and "safety" should mean in the AI era.

The November Convening was a milestone on the road to the February 2025 AI Action Summit in France, and was held on the eve of the Convening of the International Network of AI Safety Institutes. Over 45 AI experts and practitioners gathered to advance a practical, solutions-oriented approach to AI safety where two key dynamics emerged. First, while the open source AI ecosystem continues to gain traction, there is a pressing need for more open and interoperable tools to support responsible and trustworthy AI deployments. Second, this community seeks to approach safety systems and tools differently – prioritizing decentralization, pluralism, cultural and linguistic diversity, and an emphasis on transparency and auditability. The resulting collaborative output ("[A Research Agenda for a Different AI Safety](#)") informed relevant parts of the French Government's AI Action Summit.

Since the second Columbia Convening, the AI landscape has evolved from technical, governance, and funding standpoints. The report below was updated in April 2025 to reflect these developments, and its authors remain committed to the main findings of the report and the need for continued innovation in AI safety in general.

We are grateful to the twelve working group members who collaborated over six weeks to produce a 40-page Backgrounder, which laid the foundation for this paper. We also thank the broader group of participants who contributed critical research, insights, and practical solutions, and who are recognized as co-authors of this work. Lastly, we extend our sincere thanks to Mozilla and Columbia for funding and hosting this effort, and to all those who took part in the first Columbia Convening for their commitment to strengthening the open-source AI community.

## Introduction

The open ecosystem in AI is gaining momentum among practitioners and developers, with open models now spanning a wide range of modalities and sizes and performing nearly on par with the leading closed models, making them viable for most AI use cases.[1] Hugging Face reported a 880% increase in the number of Generative AI model repositories in two years, from 160,000 in January 2023 to 1.57M in November 2024.[2] According to a 2024 study by investment firm a16z,[3] 46% of Fortune 500 company leaders reported strongly preferring to use open source models. Alongside this increasing adoption of open models, many researchers, policymakers, and companies are starting to embrace model openness[4] as a benefit to safety, rather than a risk. There is growing recognition that safety is as much - if not more of - a system property than a model property.[5] This underscores the need to expand open safety research and tooling to address risks throughout the entire AI development lifecycle.

The technical and research communities invested in openness in AI systems have been developing tools to make AI safer for years, ranging from better evaluations and benchmarks, to improved documentation. Much of this work has been conducted publicly, upholding the principles of openness and embracing diverse perspectives on what safety means and how it can be effectively achieved. Accelerating openness in AI safety offers clear benefits, as AI system and model developers increasingly need access to the knowledge, tools, and safeguards necessary to protect users and society from unintended risks.

Amid these developments, disconnects persist between the related fields of Trust and Safety and Responsible AI where differences in terminology, harm and risk frameworks, and persistent organizational, educational, and cultural silos hinder the sharing of best practices, tools, and insights that could benefit the AI openness community. This paper aims to bring greater clarity and actionability to these research needs, while intentionally integrating interdisciplinary perspectives from adjacent domains and areas of expertise.

## The Columbia Convening on AI Openness and Safety

On Nov. 19, 2024, Mozilla and Columbia University's Institute of Global Politics held the Columbia Convening on AI Openness and Safety in San Francisco. The convening brought together over 45 experts and practitioners in AI to advance practical approaches to AI safety that embody the values of openness, transparency, community-centeredness, and pragmatism both in its research focus and proposed outcomes, and in how the work was conducted. After a subgroup developed a backgrounder document to frame the conversation, conveners met in person to explore how to empower AI systems developers in determining the most relevant technical interventions and associated tooling. Participants focused on mapping harms to specific interventions, highlighting gaps in safety tooling to inform community tool building priorities, and identifying pain points and barriers to adoption in safety tooling. Conveners also reflected on the methods by which openness in AI safety can foster more participatory, community-informed, context-appropriate, and diverse approaches to AI safety issues. The outcomes are reflected in this paper as follows:

1. A community-informed research agenda at the intersection of safety and open source AI to inform the February 2025 [AI Action Summit](https://www.elysee.fr/en/sommet-pour-l-action-sur-l-ia)[6] (See Section 1.4 Collaborative Research Roadmap)
2. Identification of existing and needed technical interventions and open source tools to safely and responsibly deploy open foundation models across the AI development workflow (See Section II: Mapping Post-Training Technical interventions and Tooling for Safety)
3. Mapping the content safety filter ecosystem with a proposed roadmap for future research and development (See Section 4.3-Overview of Content Safety Filters For Open Models)

---

[1] Labonne, M. (2024, July 24). "I made the closed-source vs. open-weight models figure for this moment." X. Link.
[2] Fahlgren, C. (2024, September 26). "Other platforms like GitHub also reported strong growth with a 248% year-over-year increase in the number of Generative AI model repositories in 2023." X. Link.
[3] 16 Changes to the Way Enterprises Are Building and Buying Generative AI, Wang, S., Xu, S., 2024. Link.
[4] Joint Statement on AI Safety and Openness. (n.d.). Mozilla. https://open.mozilla.org/letter/signatures/
[5] Narayanan, A., & Kapoor, S. (2024, March 12). AI safety is not a model property. AI Snake Oil. https://www.aisnakeoil.com/p/ai-safety-is-not-a-model-property

This paper expands on the key takeaways from the working group discussions and in-person workshop.

# I) Research Roadmap on AI Openness and Safety

This section reviews the scope of our research, highlighting omissions in current risk frameworks and subsequent consequences for AI safety research. A literature of risk taxonomies supporting this section is included in [Appendix 1](#).

---

[6] Artificial Intelligence Action Summit. (n.d.). elysee.fr. https://www.elysee.fr/en/sommet-pour-l-action-sur-l-ia

## 1.1 How AI Openness Contributes to Safety

Safety risks posed by foundation models to individuals, organizations, and society are central to ongoing debates about whether such models should be open, underscoring the inherent connection between safety and openness.[7] [8] While early discussions on how AI openness relates to safety were hindered by the absence of a clear definition of open source AI,[9] this community proposed a practical framework for openness in AI.[10] The Open Source Initiative's recent definition of open source AI has further clarified the boundaries of this evolving landscape.[11] Our work seeks to advance the conversation by examining how open source AI systems and tools can enhance AI safety through new participatory methods and by fostering a robust open ecosystem (See Sections 1.1 and V). We also explore how the open source community can help practitioners build safer applications using open models by identifying research gaps, technical interventions, and tooling needs (see Sections 1.4 and II).

The contributions of open models and open tooling to AI safety are a natural extension of the decades-long debate over open-sourcing dual-use cybersecurity tools. As noted by CISA, "*the general consensus among the security community is that the benefits of open sourcing security tools for defenders outweigh the harms that might be leveraged by adversaries – who, in many cases, will get their hands on tools whether or not they are open sourced. While we cannot anticipate all the potential use cases of AI, lessons from cybersecurity history indicate that we can stand to benefit from dual-use open source tools.*"[12]

For AI models, the degree of openness in different parts of the model stack[13] can enable greater transparency, scrutiny, and insight into the model's inner workings. This, in turn, can support safety improvements and more effective control over the model's outputs.[14] Initiatives like GemmaScope,[15] to open Sparse Autoencoders,[16] as one example, can significantly benefit AI safety research in the long term. Even for partially open models, where internal layers are not accessible except to the model builder, access to the logits can be helpful for safety evaluation. For instance, computing average precision and traditional AUCPR metrics is not feasible for black-box APIs like Azure API and GPT-4, as the APIs do not provide the probability scores required for this metric to the developers.

---

[7] Simonite, T. (2023) "Open-Source AI: Pros and Cons." *IEEE Spectrum*. Link. or Metz, C. (2023) "Should AI Be Open Source? Behind the Tweetstorm Over Its Dangers." *Wall Street Journal*. Link.
[8] The terms risks, harms, and hazards are often used interchangeably, but are conceptually distinct. "Risks" refer to the potential for negative outcomes (i.e., typically incorporate notions of likelihood and severity of impact), and typically have associated controls. "Hazards" refer to the sources or causes of potential harms, and "harms" are the downstream undesired outcomes. These distinctions matter when considering intentionality and responsibility in governance.
(https://www.trailofbits.com/documents/Toward_comprehensive_risk_assessments.pdf)
[9] Gent, E. (2024) "The Tech Industry Can't Agree on What Open-Source AI Means. That's a Problem." *MIT Technology Review*. Link.
[10] Basdevant, A., François, C., Storchan, V., Bankston, K., Bdeir, A., Behlendorf, B., … & Tunney, J. (2024). Towards a Framework for Openness in Foundation Models: Proceedings from the Columbia Convening on Openness in Artificial Intelligence. arXiv preprint arXiv:2405.15802. Link. Additionally, the official OSI definition on open source AI completed the clarification effort.
[11] The Open Source AI Definition – 1.0, Open Source Initiative, 2024. Link.
[12] Cable, J., Black, A. (2024) "With Open Source Artificial Intelligence, Don't Forget the Lessons of Open Source Software." *Cybersecurity and Infrastructure Security Agency (CISA)*. Link.
[13] See the proceedings of the first Columbia Convening, "Towards a Framework for Openness in Foundation Models: Proceedings from the Columbia Convening on Openness in Artificial Intelligence, https://arxiv.org/abs/2405.15802
[14] Representation Engineering (RepE) is an example of interpretability methods leading to safety improvements. Using RepE has proven to drastically decrease safety issues like hallucination (e.g. leading to +18% improvement and SOTA on TruthfulQA, the reference benchmark for truthfulness).
[15] Lieberum, T., Rajamanoharan, S., Conmy, A., Smith, L., Sonnerat, N., Varma, V., Kramár, J., Dragan, A., Shah, R., & Nanda, N. (2024, August 9). Gemma Scope: Open Sparse Autoencoders Everywhere All At Once on Gemma 2. arXiv.org. https://arxiv.org/abs/2408.05147
[16] Sparse autoencoders (SAEs) are an unsupervised method for learning a sparse decomposition of a neural network's latent representations into seemingly interpretable features. Link.

An open ecosystem also allows developers working with AI systems to maintain full control over safety tools across the entire stack which helps mitigate risks associated with unexpected model updates.

## 1.2 The Role of AI Systems Developers in Deploying AI Safely

Our research intentionally included the perspectives of AI system builders along with those of researchers and technical community groups involved in deploying open systems to ground what are often theoretical policy conversations in practicality and actionability. AI system developers are often the final link in translating safety best practices and policies into real-world operations, making their involvement critical to identifying current obstacles to strengthening AI safety in open environments.

AI system developers reportedly face a range of challenges in safely deploying open models, some of which have been identified by the AI Alliance:[17]

1. **Lack of standardization increases development time and thus cost of deploying safety methodologies.** The AI safety tooling ecosystem is nascent but growing rapidly, which has led to tool duplication and frequent lack of interoperability between tools. This issue particularly affects developers using open models, as most closed models served via APIs include some built-in safety mechanisms such as content safety filters for input and output or prompt-rewriting.

2. **The rapid pace of AI development and constant emergence of new use cases and risks makes it difficult to keep safety interventions and tooling up to date**. For example, new speech-to-text tools have been increasingly adopted in the medical field to transcribe patients' consultation records. Recent research evidenced problematic hallucinations present in the electronic health records, including racial commentaries, violent rhetoric, and even imagined medical treatments.[18]

3. **Regulatory requirements** such as the EU AI Act,[19] existing regulations on user generated content like Child Sexual Abuse Material (CSAM), and copyright law can also require technical interventions to control the inputs and outputs of AI systems.

---

[17] https://thealliance.ai/blog/the-state-of-open-source-trust
[18] McCoy, Liam G., Arjun K. Manrai, and Adam Rodman. (2024) "Large Language Models and the Degradation of the Medical Record." https://www.nejm.org/doi/full/10.1056/NEJMp2405999.
[19] EU Artificial Intelligence Act | Up-to-date developments and analyses of the EU AI Act. (n.d.). https://artificialintelligenceact.eu/

Creating the appropriate incentives and constraints for developers to prioritize safety is essential. In the absence of stable regulatory frameworks, or in contexts where regulation is still evolving, reputational risk often remains a key motivator for implementing safety measures. While regulation can help enforce standards and drive industry-wide adoption of best practices, reducing barriers to implementing technical safety interventions can be another powerful lever.

## 1.3 Scope of Work

### 1.3.1 Components of the AI Stack in Scope

This paper focuses on openness writ large, which encompasses AI safety tooling and AI systems built with models of a varying degree of openness. This paper will refer to systems and communities as *open*, rather than *open source*, to highlight the larger scope of the discussion that encompasses models for which the whole stack is not open (e.g., "open weight" models). We address both open tools and open models.

In practice, developers often use both open and closed tools.  To account for this, mappings also refer to closed tools that can be used for open models' safety (e.g., Mistral moderation API)[20], or open tools that can be used for closed models, like PAPILLON.[21]

While we focus on open AI systems—not just open models—the underlying safety of the base model remains crucial. Selecting the right LLM is the first step in a developer's journey to building an application. Although slightly outside our primary scope, the Appendix includes an overview of various existing safety-focused benchmarks and leaderboards available for foundation models at the time of publication. The authors strongly recommend that developers consult these resources when selecting models; they also highlight the lack of safety-focused multimodal benchmarks and leaderboards as a clear gap in the current ecosystem.

### 1.3.2 Risks in Scope

There are many taxonomies originating from academic research, governmental policies, and industry operationalizations that identify and categorize risks posed from AI systems.  These taxonomies often cover:

---

[20] Moderation | Mistral AI Large Language Models. (n.d.). https://docs.mistral.ai/capabilities/guardrailing/
[21] Note: PAPILLON is a system that uses local and open LLMs to create privacy-preserving  LLM queries to use closed LLMs without sending private data.

1. The risk domain (e.g., one paper consolidates many existing taxonomies across domains into top-level categories of: societal, content safety,[22] legal and rights, and systems and operational risks),[23]
2. Accountability and causality (i.e., what is the intent, which/whose actions, or what contexts, led to the increased risk and/or manifested harm),[24] and
3. The consequences, or realized harm, and associated scope and timeline (i.e., who or what has been impacted and how, including but not limited to individuals, organizations, or systems).[25]

*Intent in scope: system and user capabilities*

Risk can be distinguished based on the intentions of both the AI system's builder and its users. A system intentionally designed and tuned for harmful capabilities is akin to a weapon: regardless of context, it is likely to cause harm. In contrast, a system designed as a general-purpose tool, like a hammer, can still be misused by users: it can be weaponized deliberately, or cause accidental harm through carelessness or other factors. Unlike systems, assessing user intent can be more complex, as it is often difficult to discern or may shift over time.[26] Therefore, a user's level of ability becomes another important factor in understanding the potential for harm.

"Intentional" harms at the system level cover malintent from the builder or the user. E.g., the deliberate creation of a system to generate harm, like PoisonGPT to spread misinformation at scale,[27] or the intentional misuse of an AI system capabilities for malicious purposes, like bioterrorism.[28] "Unintentional" harms refer to cases where an AI system—or tools built without harmful intent—cause unintended negative outcomes, including situations where users unknowingly misuse the system or operate outside its intended scope. "Inappropriate deployments" refers to harms caused by deployments of immature AI systems, or deployments where system shortcomings impact end users due to misunderstandings or miscommunications around actual model behavior. Although this often arises from a lack of due diligence in defining capabilities and limitations, this involves no specific abuse per se of these systems. This is an instance of unintentional harm that is worth calling out in the scoping conversation as the responsibility lies both with the builder and the user.

---

[22] "Content safety" in this context is reminiscent of, but does not fully encompass, taxonomies used in the adjacent field of Trust and Safety (T&S), particularly as they relate to content moderation and governance of online speech. T&S as a discipline typically focuses on protecting users from harmful actors, behaviors, and content, as well as the platform design decisions and elements representing user agency and controls. AI risk frameworks add the dimension of protecting both users from AI systems, and AI systems from malicious users (the latter of whom may have a range of capabilities and levels of sophistication).
[23] Zeng, Yi, Kevin Klyman, et al. (2024) "AI Risk Categorization Decoded (AIR 2024): From Government Regulations to Corporate Policies." Link.
[24] Slattery et al. (2024) "The AI Risk Repository: A Comprehensive Meta-Review, Database, and Taxonomy of Risks From Artificial Intelligence." Link.
[25] Vidgen, Bertie, et al. (2024) "Introducing v0.5 of the AI Safety Benchmark from MLCommons." Link.
[26] Gunaratne, C. et al. (2022) "Evolution of Intent and Social Influence Networks and Their Significance in Detecting COVID-19 Disinformation Actors on Social Media". Link.
[27] https://blog.mithrilsecurity.io/poisongpt-how-we-hid-a-lobotomized-llm-on-hugging-face-to-spread-fake-news/?. Note that other such popular supposedly harmful models were actually more 'scam'. It is the case of FraudGPT and WormGPT that pretend to help to facilitate cyber attacks but were reported to claim capabilities it didn't have as illustrated: https://x.com/aviskowron/status/1685763930948255744?s=46
[28] Hendrycks, Dan, Mantas Mazeika, and Thomas Woodside. (2023) "An Overview of Catastrophic AI Risks." Link.

To date, the debate on AI openness is mostly focused on intentional harms caused by both builders and users with malintent. However, these discussions often overlook harms arising from casual system abuse, inappropriate deployments, and the actions of well-intentioned developers —factors that account for many of today's most pressing AI risks.[29]

This paper primarily focused on existing concerns from developers building AI systems with open models, which covers unintentional harms, as well as the intentional misuse of a system by low-capability users (see Risks in Scope, below). While this focus left important risks and known harms unaddressed, the authors integrated omitted risks and tradeoffs when creating the broader research agenda and intend to continue this work in further workshops. The authors acknowledge that the landscape of accountability and harm redressal in the context of these different risk scenarios is a critical component of the discussion, but exploring these aspects in depth is out of scope for this paper.

---

[29] Weidinger, L., et al., 2023 reported that the coverage of ethical and social risk evaluation overall is low and many representations of harm are poorly covered by "discriminatory bias" benchmarks like age, religion, nationality, or social class). Link.

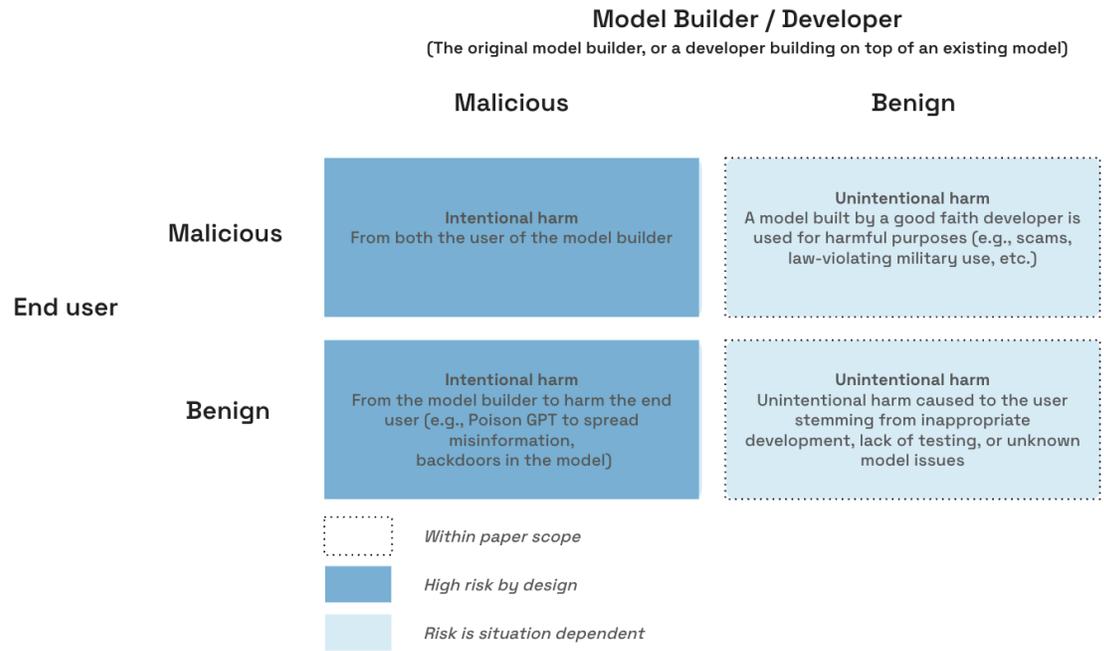

*Risk Domains in Scope*

Our work does not seek to develop a new taxonomy of harms and risks. Instead, we leverage existing taxonomies to frame our discussion and analysis, focusing on the impacts and tradeoffs of current limitations and gaps. Authors acknowledge that existing taxonomies are biased and likely overrepresent the perspectives of AI system developers. Given our goals, we decided to use existing taxonomies as a starting point to consider questions around their implementation. Aspects of AI Safety that many of these taxonomies omit are highlighted in Sections 1.3.3 and 1.3.4.

Current taxonomies of harms and safety definitions also often overlook deployment-specific and community-specific nuances due to their high level of abstraction and generalization. Just as our earlier research[30] helped advance more nuanced and granular definitions of AI openness, similarly pluralistic approaches are essential for addressing safety - particularly in areas like content moderation. Incorporating participatory feedback and community inputs into AI systems is critical for developing more context-aware safety measures. Part V explores how these participatory methods can help establish plural, context-specific and inclusive safety safeguards.

---

[30] Basdevant, A., François, C., Storchan, V., Bankston, K., Bdeir, A., Behlendorf, B., ... & Tunney, J. (2024). Towards a Framework for Openness in Foundation Models: Proceedings from the Columbia Convening on Openness in Artificial Intelligence. arXiv preprint arXiv:2405.15802. Link.

Based on a literature review of the most recent and common taxonomies ([Appendix I](#)), we consolidated public and private taxonomies of AI risk to be addressed: the [AI Risks Decoded 2024 taxonomy](#)[31], [Thorn/ATIH](#)[32], [MLCommons 0.5 AIR 2024](#)[33], [Google DeepMind (GenAI specific)](#)[34], [NIST AI 100-2e2023, NIST SP1270](#), [Weidinger, et al.](#) During the course of writing this paper, new risk frameworks emerged, including OpenAI's [Preparedness Framework Version 2](#), and an update of [Google DeepMind's Frontier Safety Framework](#). Anthropic also added minor updates to its [Responsible Scaling Policy](#).

| Risk categories | Description<br>Including any elements in and out of scope |
|---|---|
| **Child safety**<br>● [Thorn/ATIH](#)<br>● [AIR 2024](#) | Child Harm (including but not limited to grooming, minor sexualization, and illegal content such as Child Sexual Abuse Material, or CSAM). *Note, this category is separated from other content safety issues given CSAM is illegal to possess, share, or distribute in many jurisdictions. It poses unique challenges for testing and mitigations implementation.*[35] |
| **Content safety**[36]<br>● [MLCommons 0.5](#)<br>● [AIR 2024](#)<br>● [Google DeepMind (GenAI specific)](#) | Content policies are specific to an AI system and its use cases.[37] A common example is the MLCommons AILuminate taxonomy.[38]<br><br>*Categories of content safety include but are not limited to:* Violent Crimes, Non-violent Crimes, Sex-related Crimes, Child Sexual Exploitation, Indiscriminate Weapons, Suicide & Self-Harm, Hate, Specialized Advice, Defamation. |
| **Bias / Discrimination (alternatively, Legal and Rights Related)**<br>● [NIST SP1270](#) | Generation of content and/or predictive decisions that are biased, discriminatory and/or inconsistent; related to sensitive characteristics such as race, ethnicity, gender, nationality, income, sexual orientation, ability, and political or religious belief. |
| **Information risks (Privacy infringement)**<br>● [Weidinger, et al.](#) | Leaking, generating, or correctly inferring private and personal information about individuals. |

---

[31] "AI Risk Categorization Decoded (AIR 2024): From Government Regulations to Corporate Policies." arXiv.org, 25 June 2024, www.arxiv.org/abs/2406.17864.
[32] Thorn. Safety by Design for Generative AI. 2024, info.thorn.org/hubfs/thorn-safety-by-design-for-generative-AI.pdf.
[33] ML Commons. Introducing v0.5 of the AI Safety Benchmark from MLCommons. arXiv.org, 13 May 2024, https://arxiv.org/pdf/2404.12241
[34] Google DeepMind. Generative AI Misuse: A Taxonomy of Tacticsand Insights from Real-World Data.arXiv.org, 05 June 2024, www.arxiv.org/abs/123
[35] Safety by Design for Generative AI: Preventing Child Sexual Abuse, Thorn, 2024. [Link](#).
[36] There are many content safety related taxonomies; reviewing all of them is out of scope for this exercise. Sexual content is one that can be particularly fraught from a Trust and Safety perspective, and can cause challenges for teams who seek to balance user voice and autonomy, and the intended use case of the platform(s) and associated policies and regulations. T&S teams evaluating sexual content need to be able to differentiate between consensual adult material and potentially exploitative content, while also being mindful of complying with platform policies and regulatory requirements. [This set of principles](#) for image-based sexual abuse (IBSA) can serve as a guide AI systems developers.
[37] Klyman, Kevin. (2024) "Acceptable Use Policies for Foundation Models." https://arxiv.org/pdf/2409.09041.
[38] [Link](#)

| | |
|---|---|
| **Model Integrity risks**<br>● NIST AI 100-2e2023 | **In-Scope for our work:** Basic adversarial attacks like simple jailbreaking remain a focus of our collective work as they are a common threat faced by AI systems. This guidance from NIST reviews typical attack vectors like jailbreaks and data extraction, and includes mitigations.<br><br>Some of the attack types in the NIST guidance may be out of scope, such as deliberate actions by motivated, experienced adversaries aiming to *disrupt, evade, compromise,* or *abuse* the operation of the model or its output. |

Table 1 - Risk domains in scope

### 1.3.3 Areas for Further Exploration

Participants encouraged more research on discussion on the following topics:

1. **Functional failures due to premature deployment.** In many cases, the unrestricted and under-vetted use of LLMs can cause harm.[39] Examples include models used in faulty translations at the border;[40] models leveraged in incorrect audio transcriptions[41] and text summarization[42] in healthcare settings; and the corruption of critical information ecosystems leveraged by institutions (e.g. hospital records[43] and legal filings[44]) or in common use ( search engines[45]). The impact of these types of failures often extends beyond the immediate users of the AI systems, making it more difficult to identify, scrutinize, or contest harmful outputs—thereby increasing the potential for harm. Exploring these failures and their downstream consequences, and mapping the contributing factors, is a critical area for further investigation.
2. **Pre-training interventions and related tooling**. With our focus on AI system developers rather than model builders, most concerns were limited to post-training steps, including model alignment, additional safety layers like filters, AI system evaluation and monitoring practices, and data needs. One exception to this focus was with child sexual abuse material (CSAM), as the issue of filtering out CSAM from open source image datasets routinely arises as a resource gap for the open source community given the existing barriers to the proper tooling and expertise.
3. **Broader societal impacts.** While our discussion of safety tooling focused on interventions that can be applied within current AI development practices to reduce direct harms to individuals, broader safety concerns—such as safety-critical deployments and high-level design decisions[46]—were identified as important research priorities rather than areas for immediate intervention. Issues like AI's impact on work, education, information ecosystems, and creativity, as well as concerns around misinformation, environmental sustainability, and risks to financial systems, are vital and merit further exploration, but fall outside the scope of this paper.

---

[39]Raji, I.D., Kumar, I.E., Horowitz, A., & Selbst, A.D. (2022). The Fallacy of AI Functionality. *Proceedings of the 2022 ACM Conference on Fairness, Accountability, and Transparency*.
[40] Bhuiyan, J. (2023). Lost in AI translation: Growing reliance on language apps jeopardizes some asylum applications. The Guardian, 7.
[41] Garance Burke and Hilke Schellman. (2024) "Researchers say an AI-powered transcription tool used in hospitals invents things no one ever said" Associated Press.
[42]See: National Nurses United survey finds A.I. technology degrades and undermines patient safety
[43] McCoy, L. G., Manrai, A. K., & Rodman, A. (2024). "Large Language Models and the Degradation of the Medical Record". The New England Journal of Medicin*e*.
[44] Magesh, V., Surani, F., Dahl, M., Suzgun, M., Manning, C. D., & Ho, D. E. (2024)." Hallucination-Free? Assessing the Reliability of Leading AI Legal Research Tools."
[45] Robison, K. (2024). Google promised a better search experience—now it's telling us to put glue on our pizza. *The Verge*.

4. **Information security risks resulting from highly capable attacks on AI models or systems**. Mitigating advanced cyber security risks, like data poisoning or model backdoors, is critical for AI system developers but necessitates technical interventions and tooling that merit a separate discussion. A review of these risks is included in Appendix I. Basic adversarial attacks like simple jailbreaking remain in scope because of their widespread and accessible nature.

### 1.3.4 Limitations of Existing Taxonomies of Harm and Safety Definitions

The term "safety" has acquired a multitude of definitions within AI, which vary based on the context and the community. Within the context of AI, some have defined safety as the prevention of failures due to accidents,[47] [48] while others refer to the field of alignment, aiming to steer AI systems toward specific values and goals.[49] [50] These definitions have not fully captured the broader meaning of "safety" traditionally used in other fields, such as safety-critical systems like healthcare, energy, and national security. Given the nearly ubiquitous deployment of AI across all sectors and fields, a broad definition of safety is the absence of harm to people or the environment resulting from a system's outcomes.[51] [52]

Ensuring the safety of AI across different applications requires defining acceptable risk thresholds that reflect the types and severity of potential harms specific to each sector. Put another way - safety assurance of AI is not possible without considering an intended use case.[53] For example, the accepted level and range of harms in education would drastically differ from those in a healthcare setting. This also highlights why participatory approaches designed to create tools and systems tailored to specific contexts and communities matters in the context of AI safety (see below). Systems are rarely "safe" in an abstract sense.

A defined use case and corresponding risk thresholds are critical factors in guiding development decisions and determining the level of disclosure needed for effective harm reduction. Within a given use case, determining the scope of safety considerations for an AI system requires both scoping work from system developers, and meaningful integration[54] of external stakeholders' inputs. When "safety" is defined as the mitigation of risks of harm a system may cause to its environment (as opposed to "security," which addresses intentional misuse),[55] [56] understanding the scope of safety-relevant decisions and interventions requires a clear definition of the system itself, as well as the specific risks and aspects of the "environment" being considered.

---

[46] Raji, I.D., Kumar, I.E., Horowitz, A., & Selbst, A.D. (2022). The Fallacy of AI Functionality. *Proceedings of the 2022 ACM Conference on Fairness, Accountability, and Transparency.*
[47] Amodei, D., Olah, C., Steinhardt, J., Christiano, P., Schulman, J., & Mané, D. (2016). Concrete problems in AI safety. Link.
[48] Raji, I., & Dobbe, R. (2020). Concrete problems in AI safety, revisited. ICLR workshop on ML in the real world.
[49] Langosco, Lauro Langosco Di; Koch, Jack; Sharkey, Lee D; Pfau, Jacob; Krueger, David (2022). "Goal misgeneralization in deep reinforcement learning." International Conference on Machine Learning. Vol. 162. PMLR. pp. 12004–12019.
[50] Brown, D. S., Schneider, J., Dragan, A. D., & Niekum, S. (2021). Value alignment verification.http://arxiv.org/abs/2012.01557
[51] Khlaaf, Heidy. (2023) "Toward Comprehensive Risk Assessments and Assurance of AI-Based Systems." Trail of Bits. https://www.trailofbits.com/documents/Toward_comprehensive_risk_assessments.pdf.
[52] The Trust and Safety field has similarly grappled with definitions of "safety" given it is a highly contextualized concept, particularly relating to the online-to-offline spectrum. The T&S discipline has evolved from a predominant focus and reliance on reactive content moderation to be more proactive and design-oriented around harm prevention, enabling more positive outcomes, and the overall systems dynamic of a platform. The increasingly stringent regulatory environment is influencing the field in new directions as well (see Link)
[53] Khlaaf. (2023) "Toward Comprehensive Risk Assessments."

While these dimensions are difficult to generalize across the wide range of AI applications, existing approaches like those in car and road safety can offer valuable lessons and highlight the spectrum of choices that influence how safety is defined and implemented. For example, car safety manufacturing standards may focus either primarily on the safety of the driver,[57] or ensure standards incorporate the needs of specific passengers such as small children or people with disabilities. Standards may make additional requirements to protect pedestrians.[58] Safety-relevant characteristics such as the overall size of vehicles[59] highlight the tensions that exist between different stakeholder groups; in this instance, design decisions that appeal to one customer base may increase risks for other road users. Questions as to whether air pollution should be a matter of car safety[60] illuminate the complexity of unequivocally scoping safety considerations for a particular system. Thus, each application of AI raises questions over the prioritized categories of risks for categories of stakeholders (the "environment"), and which aspects of the AI component's design choices and properties constitute risk factors for identified harms (the "system").

Many discussions on AI safety disproportionately focus on direct users and viewers of an AI system's outputs, and not on a broader view of the algorithmic subjects whose lives are impacted by a system's decisions. Authors emphasized the need to broaden the dialogue around harms and safety beyond categories like bias, fairness, and harmful outputs, and include substantial harms that can arise from areas such as climate change, military applications and warfare,[61] means-testing algorithms,[62] and policing. For example, considering:

- Significant reliance on the scale of datasets and models as a way to increase model performance raises known trade-offs between commercial interests of developers and

---

[54] Arnstein, S. R. (1969). A Ladder Of Citizen Participation. *Journal of the American Institute of Planners*, *35*(4), 216–224. https://doi.org/10.1080/01944366908977225
[55] Siwar Kriaa, Ludovic Pietre-Cambacedes, Marc Bouissou, Yoran Halgand (2015), "A survey of approaches combining safety and security for industrial control systems", Reliability Engineering & System Safety, Volume 139, pp. 156-178.
[56] Khlaaf. (2023) "Toward Comprehensive Risk Assessments."
[57] As in the United States, although recent regulatory proposals have tried to bring the safety of pedestrians more directly within the purview of manufacturing standards: Federal Motor Vehicle Safety Standards; Pedestrian Head Protection, Global Technical Regulation No. 9; Incorporation by Reference
[58] Protection of pedestrians and vulnerable road users | EUR-Lex
[59] https://www.npr.org/2024/08/23/nx-s1-5084276/pedestrian-protection-bill-bigger-cars-trucks
[60] Lutz Sager (2019), "Estimating the effect of air pollution on road safety using atmospheric temperature inversions", Journal of Environmental Economics and Management, Volume 98.
[61] Khlaaf, Heidy, Sarah Myers West, and Meredith Whittaker. (2024) "Mind the Gap: Foundation Models and the Covert Proliferation of Military Intelligence, Surveillance, and Targeting." https://arxiv.org/abs/2410.14831.
[62] https://www.hrw.org/news/2024/01/05/algorithms-too-few-people-are-talking-about

- safety considerations for external stakeholders.[63] [64] This includes concerns around privacy, discrimination, and environmental harms such as contributing to climate change and negative environmental impacts on communities near data centers.[65]
- Concerns around current and anticipated uses of commercial foundation models in national security contexts, raising questions about oversight, accountability, and the potential for unintended consequences.[66] [67] This includes the inability to prevent personally identifiable information from being used in the intelligence, surveillance, target acquisition, and reconnaissance (ISTAR) capabilities of commercial foundation models. Such use may contribute to the development and spread of military AI technologies, which can create life-or-death risks for civilians, and increase the likelihood of failures that could trigger geopolitical tensions or military escalation.[68]

There are also concerns about the fitness-for-purpose of AI systems and the risks associated with deploying immature or insufficiently evaluated models in new contexts—risks that have been identified particularly by stakeholders who are most affected by safety impacts but are not part of the development process.[69] [70] In a national security and warfare context, many AI companies including Meta, Anthropic, OpenAI, Scale AI, and others have explicitly stated their willingness to make their AI products available for US national security matters, including ISTAR systems. (Recent uses of ISTAR systems have facilitated significant present-day harms through the fallible collection and use of personal information (e.g., Gospel,[71] Lavender, and Where's Daddy[72] have caused a significant civilian death toll in Gaza)). Although these systems are not foundation models, they have set a precedent for error-prone AI predictions that can cause serious harm. This pattern of risk continues with foundation models, which are now being suggested for applications such as assisting "the Pentagon computers better 'see' conditions on the battlefield, a particular boon for finding—and annihilating—targets".[73] Entities using foundation models for ISTAR applications have yet to demonstrate rigorous development approaches to substantiate assertions concerning the safety and fitness of these AI systems within military contexts.[74] Furthermore, the safety conversation has so far neglected the downstream impacts of such uses on civil society, where AI primarily developed for national security matters can be weaponized against domestic citizens.[75]

---

[63] Khlaaf, West, and Whittaker. (2024) "Mind the Gap."
[64] Abeba Birhane, Sepehr Dehdashtian, Vinay Prabhu, and Vishnu Boddeti. 2024. The Dark Side of Dataset Scaling: Evaluating Racial Classification in Multimodal Models. In Proceedings of the 2024 ACM Conference on Fairness, Accountability, and Transparency (FAccT '24). Association for Computing Machinery, New York, NY, USA, 1229–1244. https://doi.org/10.1145/3630106.3658968
[65] Urquieta, C., & Dib, D. (2024) "U.S. Tech Giants Are Building Dozens of Data Centers in Chile. Locals Are Fighting Back." *Rest of World*. Link.
[66] Wiggers, K. (2024) "Meta says it's making its Llama models available for US national security applications." *TechCrunch*. Link.
[67] Wiggers, K. (2024) "Anthropic teams up with Palantir and AWS to sell its AI to defense customers." *TechCrunch*. Link.
[68] Khlaaf, West, and Whittaker. (2024) "Mind the Gap."
[69] Raji, I.D., Kumar, I.E., Horowitz, A., & Selbst, A.D. (2022). The Fallacy of AI Functionality. *Proceedings of the 2022 ACM Conference on Fairness, Accountability, and Transparency*.
[70] e.g., risks to the patients and healthcare workers affected when hospitals adopt AI systems, see: National Nurses United survey finds A.I. technology degrades and undermines patient safety
[71] Abraham, Y. (2023) "'A Mass Assassination Factory': Inside Israel's Calculated Bombing of Gaza." *+972 Magazine*. Link.
[72] Abraham, Y. (2024) "'Lavender': The AI Machine Directing Israel's Bombing Spree in Gaza." *+972 Magazine*. Link.
[73] Biddle, S. (2024) "Microsoft Pitched OpenAI's DALL-E as Battlefield Tool for U.S. Military." *The Intercept*. Link.
[74] Khlaaf, West, and Whittaker. (2024) "Mind the Gap."

In contrast to the broad safety considerations raised in this paper, current AI safety efforts by developers working within fixed technical paradigms typically focus on narrow interventions that avoid major changes to the structure of the development pipeline. One example of such interventions is content filtering—either at the level of training data or model outputs (see Section IV)—to shape a model's behavior so that it aligns with intended use cases and is more resistant to misuse in out-of-scope contexts. This can be reinforced through additional training techniques (alignment), as well as through targeted information security measures, such as mitigating risks of data leakage between inputs, training data, and outputs, or preventing jailbreaking. The primary advantage of these interventions is that they align with established AI development practices and can often be implemented independently, without requiring significant changes to the overall development pipeline.

While this may be a reasonable starting point for those aiming to mitigate narrow risks quickly, such interventions are poorly equipped to address many of the safety gaps outlined above-particularly the tension between developers' performance goals and the safety needs of external stakeholders. The current focus on individual users, rather than commercial actors deploying AI systems at scale,[76] overlooks critical elements of development necessary for meaningful harm reduction. Relying on policy decisions alone is insufficient to address these expanded safety challenges.

To meaningfully improve the safety of AI systems for all stakeholders, it is essential to broaden both the scope of safety interventions and the range of harms and harmed parties that are explicitly recognized and addressed in this work. For example, established practice for safety-critical systems necessitates traceability, which is the procedure of tracking and documenting all artifacts throughout development and manufacturing processes. Traceability is required to guarantee that no aspect of the development pipeline is compromised to ensure a system's security and fitness for use, including identifying human labor and data sources across the supply chain.[77] The Dataset Convening, a community research effort led by Mozilla and EleutherAI, notably helped develop tools, norms and technical best practices to responsibly curate and govern open datasets[78].

Development interventions, rather than policy decision making, play a significant role in substantiating safety. The lack of traceability within foundation models has led to novel attack vectors, including poisoning web-scale training datasets and "sleeper agents" that may intentionally or inadvertently subvert models used in mission-critical applications.[79] Traceability is therefore closely aligned with the principles of openness and safety. Regardless of whether information is readily available through an open source model or accessible through auditing mechanisms for a closed source model, traceability plays a role in transparency and disclosure to appropriate stakeholders in assuring the safety of the model at hand, and its fitness for use.

---

[75] MacColl, M., & O'Kane, S. (2024) "'Whatever You Want, Ben': Inside Ben Horowitz's Cozy Relationship with the Las Vegas Police Department." *TechCrunch*. Link.
[76] e.g. Personas section in [2404.12241] Introducing v0.5 of the AI Safety Benchmark from MLCommons
[77] Khlaaf, West, and Whittaker. (2024) "Mind the Gap."
[78] Baack, Biderman, Odrozek, Skowron, Bdeir, et al "Towards Best Practices for Open Datasets for LLM Training"

Given the relative novelty and unprecedented speed and scale of AI system deployment, making progress on these questions requires greater transparency. This includes clearer insight into intended and actual use cases, improved traceability of system components, and continued research into how design decisions at all levels impact safety, particularly through investigations led by external stakeholders.

## 1.4 Collaborative Research Roadmap

Our working group identified several areas for further research to support the safe deployment of open models in AI systems and harness the benefits of openness and pluralism in the development of safety tools and practices:

### 1.4.1 Participatory inputs in safety systems

At what points in the AI development and deployment pipelines can participatory inputs and democratic engagement enhance safety tools and systems—making them more pluralistic and better adapted to specific communities and contexts? This is addressed in Section V.

### 1.4.2 Future of content safety classifiers

As model capabilities and modalities continue to evolve, content safety classifiers will as well. There is a clear need for more controllable and adaptable classifiers that can operate across a wide range of modalities. Developing a forward-looking research agenda is essential to shape the future of content filtering in the generative AI era. This topic is addressed in Section 4.5.

### 1.4.3 Safety tooling in open AI stacks

The ecosystem of open source tools for AI safety is burgeoning, which can make it difficult for developers to navigate. Additional research is needed to map technical interventions and related tooling and to identify gaps for developers to deploy systems safely. This is addressed in Section II.

### 1.4.4. Agentic Risks

While there is considerable enthusiasm for agentic applications as a new frontier in AI, it is urgent to establish a sufficiently robust working definition and to map the specific requirements for developing safe agentic systems. This includes identifying the current gaps in tools and practices, and is addressed [Section III](#).

---

[79] Nicholas Carlini, Matthew Jagielski, Christopher A. Choquette-Choo, Daniel Paleka, Will Pearce, Hyrum Anderson, Andreas Terzis, Kurt Thomas, and Florian Tramèr. (2024). Poisoning Web-Scale Training Datasets is Practical. In 2024 IEEE Symposium on Security and Privacy (SP). IEEE, San Francisco, CA, USA, 407–425. Link.

1.4.5 What's missing from our taxonomies of harm and definitions of safety
We surface multiple taxonomies of harms, but key gaps remain. At this pivotal moment, what do notions of safety popularized by governments and big tech companies fail to capture? We address this in [Section 1.3.4](#).

# II) Mapping Post-Training Technical Interventions and Tooling for Safety

This section provides a framework to map technical interventions and related toolings for scoped harms by stage of the machine learning workflow.

Safety tooling helps ensure that safety and evaluation processes are standardized, accessible and, ideally, transparent. Tools have historically played a key role in AI accountability processes;[80] for example, the OECD currently maintains a Catalogue of Tools and Metrics for Trustworthy AI.[81] However, not all tools are created equal. Recent research revealed that many of the tools from the OECD catalog use faulty or discredited metrics and methods for assessing safety attributes such as fairness or explainability.[82] We acknowledge that while mapping and vetting useful technical safety infrastructure is a start, further work is needed to understand and develop the safety tooling required to properly deploy meaningful interventions.

The framework below supports discussions on mapping technical interventions and associated safety tooling across the post-training workflow, and surfaces gaps to guide potential investment and research. It outlines a series of post-training methods and steps commonly used for implementing safety mitigations. Note that not all steps might be required, and the post-training workflow is frequently less linear than what is represented. This framework does not include alignment methods since most alignment methods (RLHF, DPO, RLAIF, refusal training etc.) are not safety-specific and require discussion beyond our scope.

---

[80] Ojewale, V., Steed, R., Vecchione, B., Birhane, A., & Raji, I. D. (2024). "Towards AI Accountability Infrastructure: Gaps and Opportunities in AI Audit Tooling."
[81] https://oecd.ai/en/catalogue/tools
[82] Kate Kaye and Pam Dixon, (2023) "Risky Analysis: Assessing and Improving AI Governance Tools" World Privacy Forum.

## 2.1 Post-Training Technical Interventions and Tooling for Safety

The tools listed below launched prior to this paper's writing and illustrate the dynamic yet fragmented ecosystem of open AI safety tools. Columbia and Mozilla have since launched [ROOST](#), a community effort to build open, scalable, and resilient safety infrastructure. ROOST develops, maintains, and distributes open source building blocks to safeguard global users and communities.

| Hazard Description | Data for tuning & evaluation (generation, collection) | Model tuning | Online filtering (and Offline data cleaning) | Evaluation | Monitoring |
|---|---|---|---|---|---|
| **Child Safety**<br>• Thorn/ATIH<br>• AIR 2024 | [Thorn AIG-CSAM](#). Closed source dataset of prompts and configuration parameters known to generate AIG-CSAM.<br><br>[PAN12](#). Training data for identifying sexual predators in chat.<br><br>[Curated PJ.](#) Curated dataset of predatory chats from Perverted Justice. | | [Haidra Horde Safety](#) Uses CLIP+BLIP for safety features for the AI Horde.<br><br>[PDQ:](#) Open source hashing algorithm to detect known CSAM<br><br>[PhotoDNA](#): Hashing algorithm to detect known CSAM<br><br>[Cybertip API:](#) Tool to report child exploitation to NCMEC<br><br>[IWF URL List:](#) Removes URLs related to CSAM | [Thorn model hashes (closed source)](#): A dataset of hashes of models that are known to generate AIG-CSAM.<br><br>[MLCommons AI Safety Benchmark](#) (ModelBench) A proof-of-concept benchmark based on LlamaGuard classifiers to assess safety of models. | [HuggingFace Provenance and Deepfake Detection](#). Tools to detect deepfake generation.<br><br>[Stable Signature:](#) Technique for watermarking generative AI. |
| Identified gaps in technical interventions & tooling | While CSAM is included in some datasets' categories of harm, few dedicated datasets exist.<br><br>The Thorn dataset listed is not open source and requires approval from Thorn. The datasets listed are only in English. Due to laws around the possession and use of CSAM, the lack of clear guardrails and safe harbor provisions make it difficult to share datasets.<br><br>There is a literature gap on techniques for fine tuning. | Few tools were identified to fine tune models for child safety. While not specific to child safety, it is critical to fine tune with localized languages. | Existing filters consist of hash algorithms which require integration with databases of known CSAM. They require permission to access (for instance, from Microsoft in the case of photoDNA) and can be difficult to deploy at scale, as they were designed for social media use cases.<br><br>These tools also only target previously hashed CSAM, leaving a gap for novel or synthetic CSAM. | There are no open benchmarks available to evaluate a model specifically for child safety. | Few tools identified to monitor child safety harms in models.<br><br>Content provenance is one method to monitor and trace abuse. |

Table 2 - Post-training technical interventions and tooling for child safety risks

| Hazard Description | Data for tuning & evaluation (generation, collection) | Model tuning | Online filtering (and Offline data cleaning) | Evaluation | Monitoring |
|---|---|---|---|---|---|
| **Content safety**<br>- ML Commons 0.5 AIR 2024<br>- Google DeepMind (GenAI specific)<br><br>**Categories of content safety include but are not limited to:** Violent crimes, Non-violent crimes, Sex-related crimes, Child sexual exploitation, Indiscriminate weapons, Suicide & self-harm, Hate, Specialized Advice, Defamation. | Helpfulness and harmlessness data. A repository of human preference data to enable RL for guiding models towards helpful and away from harmful outputs.<br><br>VLGuard. A fine-tuning dataset for safety alignment of vision-language models (VLMs).<br><br>Aegis AI Content Safety Dataset 1.0. 11k manually annotated interactions between humans and LLMs covering content safety topics.<br><br>BeaverTails. A dataset of 60k examples of helpful and harmful data used for content moderation and RLHF. | Tampering Attack Resistance (TAR). A method for building tamper-resistant safeguards into open-weight LLMs to prevent the removal of safeguards.<br><br>CTRL. A data curation framework to mitigate jailbreaking attacks during pre-training or fine-tuning. | Llama-Guard. LLM-based input-output safeguard models geared towards Human-AI conversation use cases.<br><br>ShieldGemma. A set of instruction tuned models for evaluating the safety of text prompt input and text output responses against a set of safety policies.<br><br>Perspective API. Classifiers which score content based on issues including toxicity, sexually explicit content, and threats. Can be used as a mitigation or evaluation tool.<br><br>More filers are listed in Section IV.<br><br>Prompt-rewriting techniques can also be of help and somewhat serve as input filtering. | MLCommons AI Safety Benchmark (ModelBench). a proof-of-concept benchmark based on LlamaGuard classifiers to assess safety of models.<br><br>LLM Safety Leaderboard. A unified evaluation for LLM safety, where users can submit models for evaluation based on the evaluation platform DecodingTrust.<br><br>JailbreakEval. A collection of automated evaluations to assess jailbreak attempts.<br><br>Purit. An open automation framework to proactively find security and safety risks in their generative AI systems.<br><br>WalledEval. A testing toolkit comprised of 35 safety benchmarks covering areas such as multilingual safety, exaggerated safety, and prompt injections.<br><br>SimpleSafetyTests. Handcrafted prompts covering 5 harm areas (Suicide, self-harm and eating disorders, scams and fraud, illegal items, and physical harm, child abuse), which models should refuse to comply with.<br><br>HarmBench. A standardized evaluation framework for automated red teaming<br><br>Generative Offensive Agent Tester (GOAT): An automated agentic red teaming system that simulates plain language adversarial conversations while leveraging multiple adversarial prompting techniques to identify vulnerabilities in LLMs. | SynthID Text. A watermarking generation and detection tool to identify AI-generated content (not intended for production use). |
| **Identified gaps in technical interventions & tooling.** | Many tuning datasets or techniques exist, but lack of accessible guidance on use cases can make engagement difficult.<br><br>The lack of taxonomy standard makes it more difficult to use various dataset together as they need re-labeling, which can be costly. | | Existing filters focus on a specific subset of content safety harms.<br><br>Filters primarily focus on T2T modality, and for chatbot use cases, in English. | The existence of many related benchmarks leads to a lack of standardization.<br><br>Benchmarks can be hard to access / run for model users.<br><br>More dynamic evaluations are needed while most evaluations are currently static (single-turn dataset). | Few tools identified to monitor model safety in production.<br><br>Model should be re-evaluated regularly with safety datasets containing newly sourced issues. |

Table 3 - Post-training technical interventions and tooling for content safety risks

| Hazard Description | Data for tuning & evaluation (generation, collection) | Model tuning | Online filtering (and Offline data cleaning) | Evaluation | Monitoring |
|---|---|---|---|---|---|
| **Bias / Discrimination (alternatively, Legal and Rights Related)**<br>● NIST SP1270<br><br>Generation of content and/or predictive decisions that are biased, discriminatory and/or inconsistent; related to sensitive characteristics such as race, ethnicity, gender, nationality, income, sexual orientation, ability, and political or religious belief. | BOLD: A dataset of 23,679 English text generation prompts for bias benchmarking across five domains: profession, gender, race, religion, and political ideology.<br><br>Winogender: A dataset of sentence pairs that differ solely by the gender of one pronoun in the sentence, designed to test for the presence of gender bias in automated coreference resolution systems.<br><br>Winobias: A dataset of 3,160 sentences, for coreference resolution focused on gender bias.<br><br>CrowS-Pairs: A dataset of 1508 examples that cover stereotypes across nine types of biases such as race, religion, or age. | | | Fairness Indicators: TF infrastructure to text group-based<br><br>Generative Offensive Agent Tester (GOAT): an automated agentic red teaming system that simulates plain language adversarial conversations while leveraging multiple adversarial prompting techniques to identify vulnerabilities in LLMs. | |

Table 4 - Post-training technical interventions and tooling for bias and discrimination risks

| Hazard Description | Data for tuning & evaluation (generation, collection) | Model tuning | Online filtering (and Offline data cleaning) | Evaluation | Monitoring |
|---|---|---|---|---|---|
| **Information risks (Privacy infringement)**<br>● Weidinger, et al. | **Dataset artifacts:**<br>EnterprisePII. A dataset to identify confidential data in various business documents, such as meeting notes, commercial contracts, marketing emails, performance reviews etc.<br><br>OpenPII-300k Use cases split across education, health, and psychology, academic only licenses.<br><br>Multilingual finance PII 20 PII classes, support 7 language, finance specific, Apache 2 license. | **Model artifacts:**<br>● StarPII<br>● GLiNER<br><br>**Differential privacy training libraries:**<br>● DP-transformers<br>● Opacus | Papillon filter: Uses LLM to create privacy preserving queries for the user.<br><br>PII processing: Multilingual Named Entity Recognition and PII processor.<br><br>Presidio: Provides fast identification and anonymization modules for private entities in text and images such as credit card numbers, names, locations etc.<br><br>Scrub: Provides multiple levels of scrubbing with ML to ensure optimal | **Red teaming tools:**<br>● ProPile<br>● PII attacks taxonomy | Vulnerability tracking |
| | Bigcode PII Crowdsourced PII dataset in code, 31 programming languages.<br><br>**Generation tools:**<br><br>Gretel Navigator (closed source)<br><br>**Cleaning tools:** see Online filtering plus scrubbing tools such as<br>● Flair<br>● Spacy Presidio | | anonymization of sensitive information and safeguarding of user privacy<br><br>Octopii: uses OCR, regex lists NLP to search public-facing locations or Government ID, addresses, emails etc in images, PDFs and documents.<br><br>Gitleaks: tool for detecting secrets like passwords, API keys, and tokens in git repos. | | |
| Identified gaps in technical interventions & tooling. | The datasets listed above are not fully open source as some licenses are academic only.<br><br>Lack of recent out of the box open source tools for generating PII datasets. | Need more domain specific pre-trained models that are privacy preserving fine tuned across various domains and tasks. | | There is a lack of benchmarks or leaderboards for PII evaluation. | |

Table 5 - Post-training technical interventions and tooling for information risks (privacy infringement)

| Hazard Description | Data for tuning & evaluation (generation, collection) | Model tuning | Online filtering (and Offline data cleaning) | Evaluation | Monitoring |
|---|---|---|---|---|---|
| **Model Integrity risks**<br>● NIST AI 100-2e2023<br><br>In-Scope: Basic adversarial attacks like simple jailbreaking remain in scope as it is a common threat faced by AI systems. This guidance from NIST reviews typical attack vectors like jailbreaks and data extraction, and includes mitigations. | Dataset artifacts:<br>- JBB-Behaviors: 100 distinct misuse behaviors<br>- JasperLS/prompt-injections<br>- walledai/JailbreakHub: 1,405 jailbreak prompts out of 15,140 prompts from four platforms (Reddit, Discord, websites, and open-source datasets)<br><br>- MaliciousInstruct.txt - Princeton-SysML/Jailbreak_LLM: Contains 100 malicious instructions of 10 different malicious intents for evaluation. | HarmAug: Distills large safety guard models into a 435M-parameter model.<br><br>Circuit breakers: Prevents AI systems from generating harmful content by directly altering harmful model representations.<br><br>Refusal and adversarial training methods<br>Example: "*Aligning LLMs to Be Robust Against Prompt Injection*" Link. | Prompt Guard: A classifier model trained on a large corpus of attacks capable of detecting both explicitly malicious prompts and prompts that contain injected inputs.<br><br>Safeguard LLM:<br>Prompt-Guard-86M: Detects both explicitly malicious prompts as well as data that contains injected inputs.<br><br>protectai/deberta-v3-base-prompt-injection-v2: Detects and classifies prompt injection attacks that can manipulate language models into producing unintended outputs.<br><br>deepset/deberta-v3-base-injection: Prompt injection detection and classification<br>Programmable guardrails: though they are not specific to prompt injection and jailbreaking, | JailbreakBench: Tracks performance of attacks and defenses for various LLMs.<br><br>Red teaming resistance leaderboard: Tests models with craftily constructed prompts to uncover failure modes and vulnerabilities.<br><br>DecodingTrust: Evaluates jailbreaking prompts designed to mislead GPT models to assess model robustness of moral recognition.<br><br>RapidResponseBench: Assesses how well models adapt to and mitigate various prompt injection attacks.<br><br>Latent-jailbreak: Systematic analysis of the safety and robustness of LLMs regarding the position of explicit normal instructions, word and instruction replacements. | Vulnerability tracking: https://lve-project.org/security/<br><br>Other monitoring tools are detection filters (see online filtering) |

| | | | they can allow to filter the inputs and outputs against these attacks. Nemo Guardrails | | |
|---|---|---|---|---|---|
| Identified gaps in technical interventions & tooling. | Lack of non-English datasets. | Lack of robust mitigation techniques at training time against jailbreaking and prompt injection (and verification methods). Most methods are more research artifacts than mature out of the box tools. | | | |

Table 6 - Technical interventions and tooling for model integrity risks (jailbreaking, prompt injections, etc.)

# III) Agentic Systems

This section outlines the specific safety considerations associated with agentic AI systems and identifies key gaps in knowledge and tooling needed to build and evaluate such systems with safety in mind.

## 3.1 Definition and specific use cases for agents

Agentic applications were [considered](considered) the next frontier emerging trend at the time of writing. Since then, a number of agentic applications have appeared to include [OpenAI Agent API](OpenAI Agent API), and [Claude agents](Claude agents). Various attempts have been made to define AI agents, most of which center on the idea that an agent is a system capable of performing tasks (semi) autonomously on behalf of a user or another system, while interacting with its environment. To avoid a binary definition of what is or is not an agent, we view these systems as exhibiting agent-like properties to varying degrees. As such we primarily use the term *agentic*, but we refer to the terms interchangeably in this paper. Recent [research](research) found three clusters of factors that can determine if an AI system is more or less agentic:

- **Environment and goals**. The more complex the environment—encompassing the range of tasks and domains, diversity of stakeholders, time horizons, and potential for unexpected changes—the more agentic the AI systems operating within it tend to be.
- **User interface and supervision.** AI systems that can be instructed in natural language and act autonomously on the user's behalf are more agentic.
- **System design**. Systems that incorporate design patterns such as tool use (e.g., web search, programming) or planning strategies (e.g., reflection, subgoal decomposition) exhibit a higher degree of agentic behavior.

Agents are typically tailored to specific applications, resulting in behavior and functionality that is specialized and context-dependent. This diversity makes it especially challenging to ensure the safety of agentic AI systems, as a one-size-fits-all approach is not feasible.

## 3.2 High-Potential Near Term Use Cases for Agentic AI Systems

To ground safety conversations in the real-world challenges faced by developers of agentic systems, we created use cases as the basis for discussion. It is important to note that automated actions taken by independent systems on behalf of users are not new; smart contracts, for example, have long executed predefined actions when certain conditions are met. However, the use cases outlined below introduce new safety concerns precisely because they are built on generative AI models, which necessarily introduces greater adaptability, complexity, and unpredictability.

| Name | Category | Open Source | Goal | Simulated Environments | Safety | What kind of risks might arise |
|---|---|---|---|---|---|---|
| Agent Q | Web Agent | Yes | Improve web agents' reasoning and decision-making capabilities through a combination of:<br>- Guided Monte Carlo Tree Search (MCTS)<br>- Self-critique mechanisms<br>- Iterative fine-tuning using Direct Preference Optimization (DPO). | **WebShop:** A simulated e-commerce platform used for benchmarking where agents need to find specific products<br><br>**OpenTable:** A real-world restaurant reservation website used for testing booking capabilities | No | **Financial consequences:**<br>- Incorrect booking amounts or multiple bookings<br>- Wrong payment processing<br>- Accidental cancellations of existing reservations that may have penalties<br><br>**Autonomous decision making risks:**<br>- Limited human oversight during autonomous operations<br>- Potential for compounding errors in multi-step tasks<br>- Risk of unauthorized or inappropriate actions when dealing with sensitive data or systems<br><br>**Personal data/privacy issues:**<br>- Exposing private information in forms<br>- Using wrong personal details for bookings/registrations<br>- Accidentally sharing sensitive information in public fields<br><br>**Exploration risks:**<br>- The agent might make numerous mistakes during search/exploration phases<br>- Some mistakes might be difficult or impossible to reverse<br>- Potential for unintended interactions with critical systems |
| DSPy | Framework to optimize agent design | Yes | A framework that optimizes the weights and prompts of language models algorithmically by abstracting model pipelines as graphs. | "GSM8K - Grade school math word problems<br><br>HotPotQA - Multi-hop question answering task" | No | **Optimization Risks:**<br>- Automated prompting might produce unintended or harmful behaviors<br>- Self-improving pipelines could optimize for wrong objectives<br>- Risk of compounding errors in multi-step pipelines |
| Instructor | Code Agent | Yes | A Python library that simplifies working with structured outputs from large language models. | Python code + LLMs | No | **Injection Attacks:**<br>Structured outputs processed directly without proper sanitization may be vulnerable to injection attacks, especially when integrated with other systems or databases. |
| LangChain | Agentic framework | Yes | A framework with six open-source libraries that assists with building large language model-based applications. | | No | |
| Genie | Video-based Agent | No | A software engineering model evaluated on SWE-bench and designed to emulate the thought processes of human engineers. | "Environments based on synthetic images Environments based on photographs Environments based on sketches" | No | **Content Generation Risks:**<br>- Generation of inappropriate or harmful virtual environments<br>- Recreation of dangerous or illegal scenarios<br>- Bias in generated environments from internet video data<br><br>**Action Control Risks:**<br>- Potential for learning/replicating unsafe behaviors from videos<br>- Lack of constraints on possible actions<br>- Unpredictable behavior in generated environments |

| Name | Category | Open Source | Goal | Simulated Environments | Safety | What kind of risks might arise |
|---|---|---|---|---|---|---|
| Harvey: AI Lawyer | Legal Agent (Limited info: not sure if this is an agent or LLM) | Yes | "Access laws from all US 50 states and federal, and dive deep with on-point case law interpretations and regulatory sources.<br><br>Extract nuanced and factual case law summaries better than other applications." | Legal Cases | No | - Misleading legal advice or interpretations<br>- Incorrect responses to court orders<br>- Missing required disclosures<br>- Improper ex parte communications" |

Table 8 - Examples of AI agents and related safety risks

## 3.3 Agent-Specific Safety Challenges

Most traditional AI safety approaches were developed for non-agentic systems that only generate text or images. Agentic AI systems take real-world actions that carry potentially significant consequences. For example, an agent might autonomously book travel, file pull requests on complex code bases, or even take arbitrary actions on the web, introducing new layers of safety complexity. This difference between agentic and regular AI systems changes the diversity and the nature of harms agentic systems can cause to users or society. Some differences include:

- Safety risks in agentic systems are more difficult to anticipate because seemingly innocuous actions can lead to harmful downstream consequences in the real world. For example, a simple task like booking a hotel and a cab could pose risks to physical safety if the booking is made in an unsafe neighborhood late at night.
- Actions that are harmless on their own may become problematic when combined. For instance, individual steps involved in synthesizing a drug may be innocuous, but when executed in a specific sequence could lead to illegal activity.
- The composability of steps in a task or decision-making process results in a vast number of possible scenarios, significantly increasing the number of potential risk vectors and complicating mitigation strategies.

To help developers systematically consider the full range of potential failures in agentic systems, we created a high-level system design framework and identified risks associated with specific components.

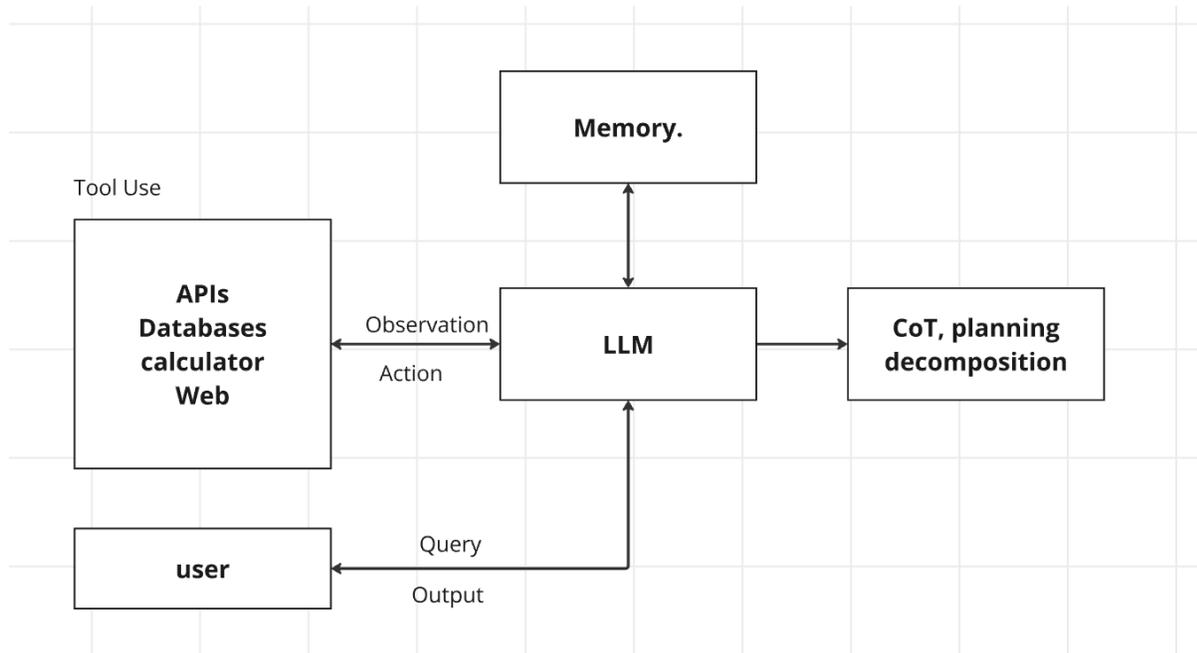

Figure 1 - High Level Agentic system Design

In this framework, a user query follows a path where (1) the user makes a direct or indirect request to the LLM, (2) the LLM analyzes the query and elicits additional preferences required to clarify the goal, (3) the agent plans the various steps to reach the desired output, (4) the agent uses specific tools to perform various steps of the journey, until it has completed the journey, and (5) achieves the goal. Each of these steps introduces risks.

## 3.4 How does safety of agentic systems differ from safety of generative models?

The fundamental differences between agentic systems and generative model-based systems require rethinking safety measures across the entire development workflow. In the context of agentic systems, evaluation benchmarks and red-teaming (simulated attacks) become especially critica

- **Evaluation Benchmarks:**[83] For agentic AI, benchmarks must assess both the accuracy of text and the safety of actions, including ethical implications and real-world impact. Unlike non-agentic benchmarks, these benchmarks need to account for the unintended

---

[83] Agent safety evaluation benchmarks:
- EICU-AC (Xiang et al., 2024): assesses privacy-related access control for healthcare agents.
- Mind2Web-SC (Xiang et al., 2024): performs safety evaluation for web agents.
- AgentHarm (Andriushchenko et al., 2024): includes a diverse set of 110 explicitly malicious agent tasks (440 with augmentations), covering 11 harm categories including fraud, cybercrime, and harassment.
- HaicoSystem (Zhou et al., 2024) a framework examining AI agent safety within diverse and complex social interactions.
- BELLS (Dorn et al., 2024): a framework for future-proof benchmarks in LLM-agents safeguard evaluation.

consequences of actions, and compliance with regulatory standards.

- **Offensive Red Teaming (Attack Simulations):**[84] Red teaming for agentic systems must extend beyond testing for harmful content generation and simulate real-world decision-making scenarios to probe for potential risks. This involves testing agents' responses to edge cases or adversarial scenarios where goal misalignment or unintended actions could lead to harm. Such offensive red-teaming with AI agents[85] can help identify risks that may require significant system changes beyond alignment or simple filtering. As a result, some developers might be more incentivized to perform simpler red teaming, whose findings can be more easily fixed but which may not address risks meaningfully.

To address safety-related issues arising from combined actions, agentic safety mechanisms must:

1. **Assess Actions Individually:** Each action taken by an agent, such as making a booking, sending a command, or interacting with sensitive data, needs to be scrutinized for potential harm. Safety guards should evaluate each action within its context to ensure compliance with both ethical standards and user intent.
2. **Evaluate Compositional Risks:** Actions that are harmless individually may become problematic when combined. Consider a banking AI agent with three seemingly harmless abilities: checking account balances (just viewing information), creating transaction templates (saving payment details without execution), and setting up automated payment rules (each following standard limits). While each action appears safe in isolation, their combination could enable a sophisticated pattern of financial manipulation. The agent could analyze account activity to identify optimal timing, create multiple small transaction templates that individually stay under warning thresholds, and then orchestrate these transfers through carefully timed automation rules. For example, instead of one flagged $30,000 transfer, it could schedule three rapid $9,999 transfers at 3 AM when monitoring is lowest. This composition bypasses traditional security measures that only examine individual transactions, creating a coordinated movement of large sums that appears innocuous when viewed piece by piece.

Safety for agents thus requires intent detection, environmental awareness, state tracking, and assessment of a system's real-time decision-making. Instead of evaluating actions in isolation, safety for agents should use multi-turn interaction evaluations, taking into account sequences of user-agent exchanges. The goal should be to understand how actions interact across different steps and contexts, to prevent sequences that might lead to undesirable or harmful outcomes, even when each step individually appears benign.

---

[84] AgentDojo (Debenedetti et al., 2024): A dynamic environment to evaluate attacks and defenses for LLM agents.
InjecAgent (Zhan et al., 2024): Benchmarking Indirect Prompt Injections in Tool-Integrated Large Language Model Agent.
[85] Narayanan, Arvind, and Sayash Kapoor. (2024) "AI Safety Is Not a Model Property."
https://www.aisnakeoil.com/p/ai-safety-is-not-a-model-property.

To map gaps in agentic-specific safety knowledge and tooling, and highlight differences with traditional AI system safety, this paper suggests using the post-training workflow framework from Section II, and highlights risks that would not be covered for the specific use cases from Section 3.2.

## 3.5 Agent Safety

With these distinct safety considerations in mind, we turn to practical realities of building agentic systems. We explore not only current best practices and open questions, but also examine newly released consumer-facing agents to understand how safety mechanisms are being implemented (or not) in the real world.

Different safety strategies offer trade-offs when securing generalist web agents. For example, using classifiers to detect prompt injections can improve safety, but may also increase false positives, leading to reduced agent autonomy and degraded user experience. Still, given the high risk of prompt injection, basic defenses remain important. Other approaches include external filters on agent actions, such as limiting spending beyond a set threshold or flagging state-changing actions that trigger user approval. These strategies help prevent high-impact errors but can also interrupt workflows or restrict useful behavior, highlighting the need to balance safety, autonomy, and usability.

Beyond monitoring individual actions, there is also a need for trajectory-level classifiers that evaluate the full sequence of a web agent's behavior. This helps detect compositional attacks that may go unnoticed when analyzing steps in isolation. One approach could be to use a secondary agentic model that monitors and assesses the main agent's actions.

Since this paper's writing, several agent developers have released consumer products. OpenAI, Google, and Perplexity all released "deep research" products that can browse the web for up to 30 minutes and create detailed reports on any topic.[86] OpenAI and Anthropic also released computer-use models that can take actions on behalf of users; Anthropic released an API, whereas OpenAI launched a product named "Operator." [87]

With regard to safety considerations, for example, Convergence's Proxy came with no defenses. Anthropic developed a classifier to prevent prompt injection. OpenAI trained Operator to detect prompt injections but it requires active user supervision on sensitive websites like email. Operator also implemented an additional model to monitor and pause execution if it detects suspicious content on the screen.[88] Such defenses can be helpful in preventing safety incidents related to agents. Figure 2 depicts a range of possible agent designs and the associated tradeoffs between security and agency.

---

[86] OpenAI, Introducing Deep Research, 2024,https://openai.com/index/introducing-deep-research; Google, *Deep Research Overview*, 2024, https://gemini.google/overview/deep-research/?hl=en; Perplexity, *Introducing Perplexity Deep Research*, 2024, https://www.perplexity.ai/hub/blog/introducing-perplexity-deep-research.
[87] OpenAI. "Introducing Operator." January 23, 2025. https://openai.com/index/introducing-operator/.
[88] OpenAI, Computer-Using Agent, 2025 https://openai.com/index/computer-using-agent/

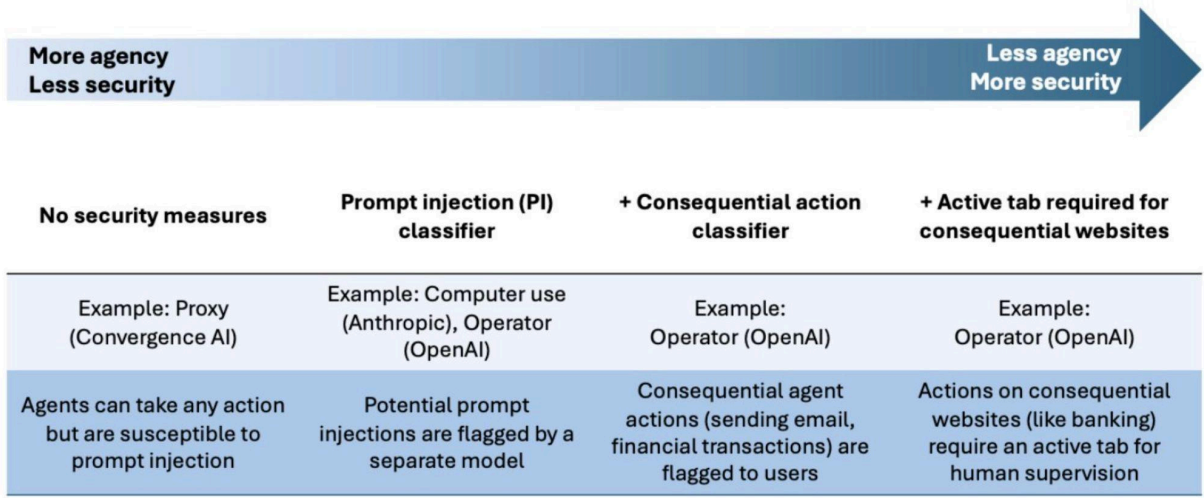

Figure 2 – The Security-Agency Tradeoff [89]

---

[89] Kapoor, Sayash (@sayashk). "Convergence's Proxy web agent is a competitor to Operator. I found that prompt injection in a single email can hand control to attackers: Proxy will summarize all your emails and send them to the attacker! Web agent designs suffer from a tradeoff between security and agency." *X (formerly Twitter)*, March 3, 2025. https://x.com/sayashk/status/1896676840103887013

## 3.6 Research Areas for Agentic Safety

The following research questions are intended to guide developers in building safer agentic systems:

1. **Taxonomy of actions, risks, and mitigations.** The developer community would benefit from a clear taxonomy of agentic actions (i.e., searching information, sharing user information, purchasing, etc. ) with their associated risks and possible mitigations.
2. **Expanding notions of consequence**. The increased complexity of the scope of safety requires redefining the current taxonomy of harms to include real-life consequences of apparently innocuous actions.
3. **Human-Computer Interaction insights.** Many real-world risks stem from misspecified user requests. More research in Human-Computer Interaction is needed to develop user-agent interaction models that reduce the likelihood of harmful real-world actions.
4. **Accountability and liability regimes**. Advancing policy research on models of stakeholders' accountability and liability is key to enforcing safety mitigations.
5. **Guardrails.** More research on guardrails for agentic systems is needed. This includes expanding beyond current filters narrowly focused on preventing specific outputs to a more holistic approach that ensures the agentic system remains aligned with its goal, to include addressing compositional risks.

# IV) Content Safety Filters

This section outlines both the opportunities and risks of using content filters in AI systems. It includes a comparative analysis of open-source content safety filters and proposes a template to help standardize their documentation. The goal is to support consistent evaluation and spark discussions on how to improve content filtering.

## 4.1 The purpose of content safety classifiers

Safety classifiers are commonly used both "online" (when the AI system is live) to filter potentially harmful content from users or the model itself, and "offline" (during pre-training, fine-tuning, or evaluation) to help curate datasets and guide model design before deployment.

### 4.1.1 Online filtering

Despite model developers' interventions to prevent risks highlighted in [Section II](), AI systems are still at risk of generating undesirable content. A common technique to prevent undesirable and potentially harmful system outputs is input and output data filtering, sometimes referred to as safeguards. Input and output filtering techniques rely on machine learning models to check if the data going into or coming out of the model complies with the policies defined for the application.

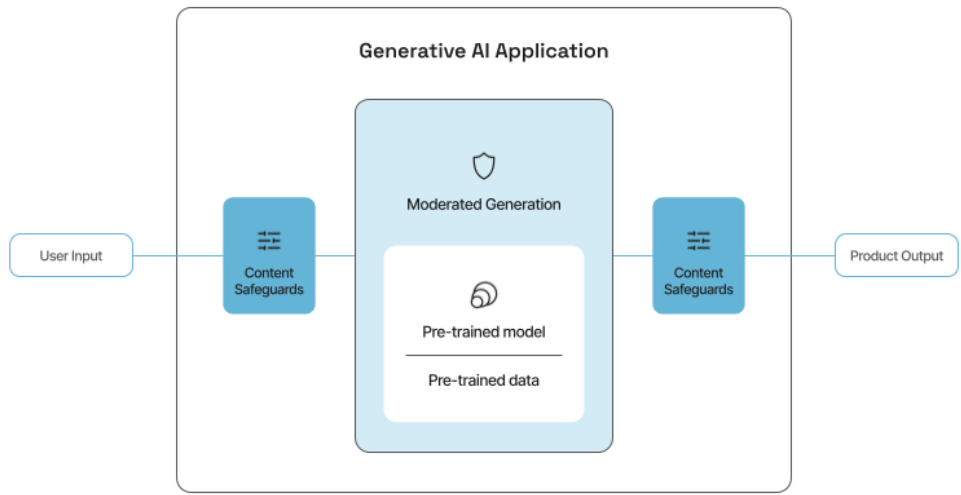

Figure 3 - Overview of an AI system with input and output filter

**Input classifiers** are typically used to filter content that is not intended to be used in the application. This could include content that violates the developer's acceptable use policy or might cause the model to produce violative outputs. Input filters often target adversarial attacks that try to circumvent acceptable use policies. **Output classifiers** are used to filter model output. These classifiers check for generated outputs that may violate safety policies.

The main difference between online and offline classifiers is latency, since low latency is necessary to guarantee a high-quality user experience when using online filtering. Online filtering may also produce metadata that is useful for offline filtering, as it gives empirical evidence of how a system is being used and how content safety classifiers may be evaded.

### 4.1.2 Offline filtering

Beyond filtering of inputs and outputs in a production system, similar types of classifiers can be used for various interventions along the AI workflow in pre-training or post-training phases. For example, offline filtering may be used in the following ways:

- Clean finetuning data by removing undesirable content such as CSAM, PII, or copyrighted material (where possible to detect).

- Create evaluation and fine tuning datasets by leveraging classifiers to up-sample adversarial prompts from user logs, synthetic datasets, or public data[90].
- Monitor model behavior and identify issues in model output logs.

Different classifiers may be required for unique functions or points throughout the AI workflow. For example, a classifier that is used in the model development pipeline will not be able to identify undesirable content if it was unable to do so in production. One way to address this is to opt for a more performant but slower model for this offline filtering step. For instance, a developer might use a small model like LlamaGuard 1B in production, and a larger one like ShieldGemma 27B offline.

---

[90] Baack, Biderman, Odrozek, Skowron, Bdeir, et al "Towards Best Practices for Open Datasets for LLM Training"

## 4.2. Content safety filters' tradeoffs

Using safety classifiers as online safeguards is a widespread practice in most chat applications, but gaps remain in this approach to system safety. Content safety filters typically make a system more complex and may add latency. They may also not be sufficiently customizable to enable them to be tailored to a specific application. Safeguards of this kind introduce tradeoffs that developers need to consider.

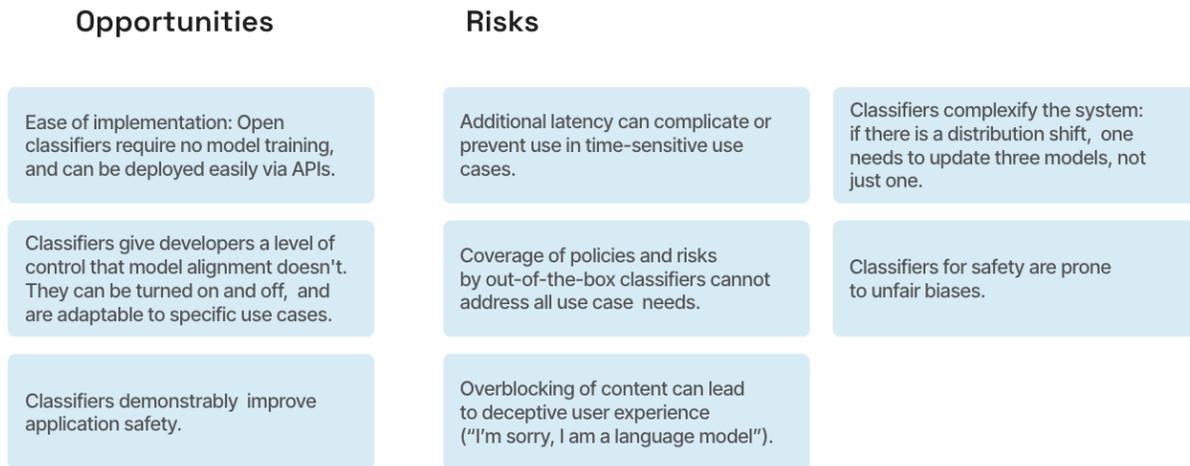

Figure 4 - Risks and opportunities of using content safety filters

Most of these tradeoffs have technical workarounds. For instance, when content is blocked, many popular chatbots handle provide a standardized refusal (e.g., some version of "I am sorry, I am a language model, I can't help you with this request"). This type of refusal can be perceived by the user as unhelpful, and in itself can be harmful if it is deployed selectively or in certain high-risk scenarios.

Key tradeoffs developers should consider when using and choosing among classifiers:

1. **Planning for unreliable performance:** Content safety systems are far from perfect and will result in mistakes, including both false positives (e.g. classifying an output as unsafe when it is not) and false negatives (failing to classify an output as unsafe when it is). Developers should evaluate classifiers with metrics like F1, Precision, Recall, and AUC-ROC to evaluate tradeoffs between false positives and false negatives. Adjusting the threshold of classifiers can help find a balance that avoids over-filtering outputs while still providing some safety benefits.
2. **Considering discriminatory biases:** Like other ML models, safety classifiers can reflect and amplify societal biases, often over-triggering on content related to marginalized identities. For example, early versions of Google's Perspective API [91] flagged comments mentioning words like "Black," "Muslim," or "gay" as more toxic, due to biased training data. This stems from imbalanced datasets where identity terms frequently appear in abusive contexts, leading classifiers to wrongly flag neutral or supportive comments. Developers must evaluate for bias (see Section 3) and apply mitigation strategies. Recent findings also show over-triggering in non-English languages like German, raising concerns about uneven moderation across linguistic contexts. [92]

3.  **Classifiers can complicate the system:** Integrating classifiers into a system introduces technical challenges, such as added latency and higher memory usage, which can be particularly problematic for devices like low-memory smartphones. Classifiers can increase the system's complexity and maintenance burden, as classifiers themselves can become sources of undesired behavior. For example, in cases of data distribution drift, both the main model and the classifiers may require retraining to remain effective.

## 4.3 Overview of content safety filters for open models

AI developers often struggle to find classifiers that meet their specific needs around behavior, latency, and performance. Community feedback points to the need for a standardized reporting framework to support comparison and informed selection. The framework introduced in this section aims to lay the foundation for that effort.

The models in this comparison vary in type—some are LLM-based, others use traditional machine learning—and differ in how and to what extent they are open. These distinctions are reflected in the framework below.

---

[91] Borkan, Daniel, et al. (2019) "Unintended Bias and Names of Frequently Targeted Groups." https://medium.com/jigsaw/unintended-bias-and-names-of-frequently-targeted-groups-8e0b81f80a23.
[92] Nogara, Gianluca, et al. (2023) "Toxic Bias: Perspective API Misreads German as More Toxic." Link.

|  | Use case and risk-coverage ||||||
|---|---|---|---|---|---|---|
|  | Filter type | Use Case | Input Modalities | Multilingual | Guarding objective | Policy coverage |
| Definition | What kind of tech artefact is used (definitions in appendix) | What risk is being mitigated? | What kind of modalities does the filter handle? | Is the classifiers supporting various languages (or code languages)? | Input vs output vs multi-turn dialogue filtering | Specific subcategories of policies that are covered by a model out of the box |
| LlamaGuard 3 8B | Safeguard LLM | - HAP Detection<br>- CSAM Detection<br>- Text Quality<br>- Privacy and anonymization | Text | 8 Languages (3) | Input filtering<br>Output filtering | S1: Violent Crimes    S7: Privacy<br>S2: Non-Violent Crimes    S8: Intellectual Property<br>S3: Sex-Related Crime    S9: Indiscriminate Weapons<br>S4: Child Sexual Exploitation    S10: Hate<br>S5: Defamation    S11: Suicide & Self-Harm<br>S6: Specialized Advice    S12: Sexual Content<br>   S13: Elections |
| LlamaGuard 3 1B | Safeguard LLM | | Text | 8 Languages (3) | Input filtering<br>Output filtering | |
| LlamaGuard 3 1B Pruned & Quantized | Safeguard LLM | | Text | 8 Languages (3) | Input filtering<br>Output filtering | |
| LlamaGuard 3 11B Vision | Safeguard LLM | | Text + Image | 8 Languages (3) | Input filtering<br>Output filtering | |
| PromptGuard | Safeguard LLM | Simple jailbreaking | Text | 8 Languages (3) | Input filtering<br>Output filtering | Jailbreaking, prompt injections |
| CodeShield | Safeguard LLM | Insecure code generation detection | Text | 7 programming languages, with an average latency of 200ms | Input filtering<br>Output filtering | Insecure code suggestion risk, secure command execution protection |
| ShieldGemma 2B | Safeguard LLM | Sexually explicit, dangerous content, hate and harassment labelling. | Text | English only | Input filtering<br>Output filtering | Sexually explicit, dangerous content, hate and harassment. |
| ShieldGemma 9B | Safeguard LLM | | Text | English only | Input filtering<br>Output filtering | |
| ShieldGemma 27B | Safeguard LLM | | Text | English only | Input filtering<br>Output filtering | |
| WildGuard | Safeguard LLM | HAP Detection<br>CSAM Detection | Text | English only | Input filtering<br>Output filtering | Malicious Uses: Fraud/Assisting illegal activities, Defamation / Encouraging unethical or unsafe actions, Mental Health crisis; Harmful Language: Social stereotypes and unfair discrimination, Violence and physical harm, Toxic language / Hate speech, Sexual content; Misinformation: Disseminating false or misleading information, Causing material harm by disseminating misinformation; Privacy: Sensitive information (Organization / Government), Private information (Individual), Copyright violations. |
| Aegis-AI-Content-Safety-LlamaGuard-LLM (NVIDIA) | Safeguard LLM | HAP Detection<br>Privacy and anonymization | Text | English only | Input filtering<br>Output filtering | Hate /Identity Hate Other: Sexual Illegal Activity, Violence Immoral/Unethical, Suicide and Self Harm Unauthorized Advice, Threat Political campaigning /Misinformation, Sexual Minor Fraud /Deception, Guns /Illegal Weapons, Copyright/trademark/plagiarism, Controlled /Regulated substances Economic Harm, Criminal Planning /Confessions High Risk Government Decision Making, PII /Privacy Malware /Security, Harassment Safe, Profanity Needs Caution |
| MD-Judge | Safeguard LLM | HAP Detection<br>Privacy and anonymization<br>Simple jailbreaking | Text | English only | Input filtering<br>Output filtering | Representation & Toxicity Harms<br>Misinformation Harms<br>Information & Safety Harms<br>Malicious Use<br>Human Autonomy & Integrity Harms<br>Socioeconomic Harms |
| Beaver-Dam-7B | Safeguard LLM | HAP Detection<br>Simple jailbreaking | Text | English only | Input filtering<br>Output filtering | Animal Abuse, Child Abuse, Controversial Topics, Politics, Discrimination, Stereotype, Injustice, Drug Abuse, Weapons, Banned Substance, Financial Crime, Property Crime, Theft, Hate Speech, Offensive Language, Misinformation Regarding ethics, laws, and safety, Non-Violent Unethical Behavior, Privacy Violation, Self-Harm, Sexually Explicit, Adult Content, Terrorism, Organized Crime, Violence, Aiding and Abetting, Incitement |
| HarmBench Classifier (Mistral-7b-val-cls) | Safeguard LLM | HAP Detection | Text | English only | Input filtering<br>Output filtering | Cybercrime & Unauthorized Intrusion, chemical & biological, weapons/drugs, copyright violations, misinformation & disinformation, harassment & bullying, illegal activities, and general harm. |
| Bespoke-MiniCheck-7B | Safeguard LLM | Factuality Hallucination detection | Text | English only | Input filtering<br>Output filtering | Fact checking based on different documents that come from diverse sources, including Wikipedia paragraphs, interviews, and web text, covering domains such as news, dialogue, science, and healthcare. |
| Llama-3-Patronus-Lynx-70B-Instruct | Programmable guardrails | Factuality Hallucination detection | Text | English only | Input filtering<br>Output filtering | Real-world domains that include finance and medicine. |
| Guardrails AI | Programmable guardrails | Factuality<br>HAP detection<br>CSAM detection<br>Text quality<br>Privacy and anonymization | Text | English only | Input filtering<br>Output filtering | Etiquette, brand risk, factuality, formatting, invalid code, jailbreaking, code exploits, data leakage. |
| Nemo Guardrails | Programmable guardrails | Privacy and anonymization | Text | English only | Input filtering<br>Output filtering | Common LLM vulnerabilities, such as jailbreaks and prompt injections |
| Guidance AI | Programmable guardrails | Factuality<br>HAP detection<br>CSAM detection<br>Text quality<br>Privacy and anonymization | Text | English only | Input filtering<br>Output filtering | |
| Pysentimiento | Non-LLM ML Guardrails | Sentiment | Text | Multilingual; ES, EN, IT, PT | Input filtering<br>Output filtering | Sentiment Analysis<br>Hate Speech Detection<br>Irony Detection<br>Emotion Analysis<br>NER & POS tagging<br>Contextualized Hate Speech Detection<br>Targeted Sentiment Analysis |

|  | Model Access | | | Technical Requirements | | | | | Documentation | Advanced Features |
| --- | --- | --- | --- | --- | --- | --- | --- | --- | --- | --- |
|  | Openness | Licence | Distribution | Performance | Size | Filter output format | Context Window | Link to source | Explainability | Customization & Control |
| Definitions | What artifacts have been released? | Indicate type of license + add link to the license | Local /on prem vs cloud-based only | Link to evaluation results - technical papers |  | Does it out put a score, a label or a text etc? | Size of the input text to be analyzed? | Add a link to the website | Does it provide a rationale for its output? | Maximum level of control and adaptability provided |
| LlamaGuard 3 8B | Weights | Llama License | API or Weights | See model card | 8B | Text | 128K | Link | No | Yes |
| LlamaGuard 3 1B | Weights | Llama License | API or Weights | See model card | 1B | Text | 128K | Link | No | Yes |
| LlamaGuard 3 1B Pruned & Quantized | Weights | Llama License | API or Weights | See model card | ~100MS | Text |  | Link | No | Yes |
| LlamaGuard 3 11B Vision | Weights | Llama License | API or Weights | See model card | 11B | Text | 128K | Link | No | Yes |
| PromptGuard | Weights | Llama License | API or Weights | See model card | 86M | Label | 512 | Link | No | Yes |
| CodeShield | N/A | MIT License | API Only | Low Latency | N/A | Label | N/A | Link | No | Yes |
| ShieldGemma 2B | Weights | Gemma License | Weights only | See model card | 2B | Text |  | Link | No | Yes |
| ShieldGemma 9B | Weights | Gemma License | Weights only | See model card | 9B | Text |  | Link | No | Yes |
| ShieldGemma 27B | Weights | Gemma License | Weights only | See model card | 27B | Text |  | Link | No | Yes |
| WildGuard | Weights Training Dataset Documentation | Apache-2.0 | Weights only | See model card | 7B | Text | 32K | Link | No | Yes |
| Aegis-AI-Content-Safety-LlamaGuard-LLM (NVIDIA) | Weights Documentation | Llama License | Weights only | See model card | 7B | Text | 4096 | Link | No | No |
| MD-Judge | Weights Training Dataset Documentation | Apache-2.0 | Weights only | See model card | 7B | Text | 32K | Link | No | No |
| Beaver-Dam-7B | Weights Training Dataset Documentation | Llama License | Weights only | See model card | 7B | Text | 2048 |  | No | No |
| HarmBench Classifier (Mistral-7b-val-cls) | Weights Training dataset Documentation | Llama Licence | Weights only | See paper | 7B | Text | 32K | Link | No | No |
| Bespoke-MiniCheck-7B | Weights Documentation | CC BY-NC 4.0 | Weights only | See model card | 7B | Text | 128K | Link | No | No |
| Llama-3-Patronus-Lynx-70B-Instruct | Weights Documentation | CC BY-NC 4.0 | Weights only | See model card | 70B | Text | 128K | Link | No | No |
| Guardrails AI | N/A | Apache 2.0 | Weights only | N/A | N/A | Text | N/A | Link | No | Yes |
| Nemo Guardrails | N/A | Apache 2.0 | Weights only | N/A | N/A | Text | N/A | Link | Yes | Yes |
| Guidance AI | N/A | MIT License | Weights only | N/A | N/A | Text | N/A | Link | Yes | Yes |
| Pysentimiento | Weights Documentation | Non Commercial | Weights only | See Paper | 125M | Text | N/A | Link | No | No |

Table 10 - Table of content safety filters

## 4.4 Current limitations and barriers to adoption

Content safety filters have various limitations:

- The lack of documented, comprehensive guidance for developers to address basic safety issues and thoughtfully implement safety filters is one of the major hurdles to adoption.
- The limited coverage of modalities, which often do not address modalities such as video, speech, or combinations of modalities, and limited languages coverage.
- Most filters are static in nature and may not adapt or generalize well across applications or domains. For example, some classifiers are trained on raw-text but then used in a conversational context. In addition to being potentially out of domain for a specific use case, they are trained on a fixed set of policies which may not always align with priority user policies given a specific use case. Even though some open source solutions support zero-shot or few-shot prompting with diverse taxonomies to generalize to novel policy dimensions, this usually requires additional work and evaluation and lacks clear guidance.
- The lack of an open and standardized evaluation mechanism across various types of content or language can lead to unknown performance issues. This requires developers to perform time-consuming evaluations themselves. The discovery of Perspective API's uneven performance across languages is a case in point: the API appears to be blocking more content in German than in other languages, requiring developers to detect such variations of performance by running their own tests.[93]

## 4.5 Vision for future developments

Our research highlights key directions for future work on classifiers, including the need for greater controllability, broader modality coverage, improved transparency, and stronger evaluation practices. Several common themes have also emerged from the developer community, including:

1. **Developing a practical "cookbook" could help guide and empower developers implementing safety filters throughout the deployment process.** Currently, a lack of clear guidance on basic safety issues creates friction. Developers face challenges in selecting relevant policies, identifying tools aligned with their product needs, and finding guidance tailored to their expertise, use case, modality, language, and jurisdiction.
2. **Larger modality coverage is needed.** As highlighted in Table 10, most content safety classifiers are currently designed for text, or text and images,[94] with other modalities or agentic systems mostly out of scope.

---

[93] Nogara, et al. (2023) "Toxic Bias."

[94] It's worth noting a helpful exception here, that of Roblox open-sourcing one of its voice safety models in July 2024: https://corp.roblox.com/newsroom/2024/07/deploying-ml-for-voice-safety

3. **Better multilingual coverage.** While some traditional ML classifiers cover a large language spectrum (Perspective API is available in 18 languages for instance[95]), most LLM safeguards are available in English only, or in a reduced set of languages as indicated in the overview Table.
4. **Developers seek greater control over filters—including the ability to adjust thresholds, customize policies, and set exceptions**. Several emerging approaches support this direction. For example, Google's agile classifiers framework[96] enables developers to create custom safeguards by tuning LLMs with just a few hundred datapoints. Further research is needed into parameter-efficient tuning methods and how to make these models production-ready.
5. **Developers want more transparency about filters**. As working group participants noted, it is often unclear why a filter blocks a specific input or output. Some approaches, such as autoraters[97] (models that provide explanations for their decisions) can offer a useful foundation for addressing this need and prompting further discussion.

Since the time of writing, the landscape of available classifiers and other content safety filters available for AI developers has changed, allowing for more opportunities for use, testing, and integration for system-level safety. As mentioned previously, the ROOST initiative aims to build open source tools for online safety, including child safety detection tools such as hash matching and review consoles for scaled labeling and reporting to NCMEC and other organizations.

# V) Pluralistic and Participatory Approaches

## 5.1 Why participation?

The authors highlight four key, non-exhaustive, reasons to incorporate collective input into AI safety efforts. We argue that no model can be universally "aligned," and that collective input and collective intelligence are essential for building more effective AI systems across diverse use cases, as well as for preventing or countering harmful applications.

1. **To surface valuable but hard-to-intuit insights and to gather a wider range of constructive ideas.** Many AI safety challenges involve complex sociotechnical systems where critical information is distributed across diverse stakeholders. Local/contextual knowledge and lived experiences of affected communities may not be captured in traditional technical documentation. Collective input helps surface edge cases, unintended consequences, and real-world failure modes that developers have been known to miss, such as via incident reporting. Tapping into the wisdom of the crowd is important, as many valuable ideas are missed when ideation is limited to small, homogenous groups.
2. **Offsetting conflicts of interest or bias.** Complex technical systems benefit from multiple independent layers of review to catch potential failure modes. Different stakeholders bring diverse priorities and risk perceptions, leading to more robust evaluations. Institutional incentives often conflict with safety goals, making external input essential for accountability. When the "default state" of a technology is unclear, collective processes offer a more grounded foundation for decision-making.
3. **Supporting normative goals around agency and input.** Democratic principles assert that those affected by AI systems should have a voice in their development and oversight. Collective input fosters legitimacy and trust by enabling meaningful stakeholder participation in safety decisions. Inclusive processes help ensure AI systems reflect diverse human values, rather than narrow technical priorities, and reinforce human agency and self-determination in shaping AI's societal impact.
4. **Bolstering a broader ecosystem.** Collective input strengthens the broader AI landscape, including public labs and independent researchers. This contributes valuable assets and helps prevent the concentration of knowledge and influence within a few large technology companies, whose investments often surpass government spending.[98]

---

[95] Perspective API is free and available to use in Arabic, Chinese, Czech, Dutch, English, French, German, Hindi, Hinglish, Indonesian, Italian, Japanese, Korean, Polish, Portuguese, Russian, Spanish, and Swedish.
[96] Mozes, Maximilian, et al. (2023) "Towards Agile Text Classifiers for Everyone." https://arxiv.org/pdf/2302.06541.
[97] Vu, Tu, Kalpesh Krishna, Salaheddin Alzubi, Chris Tar, Manaal Faruqui, and Yun-Hsuan Sung. (2024) "Foundational Autoraters: Taming Large Language Models for Better Automatic Evaluation." https://arxiv.org/pdf/2407.10817.

## 5.2 Mapping participation through the pipeline

The AI development, deployment, and post-deployment pipeline is complex, and opportunities for participation vary across stages. While we do not aim to map every possible input point, key areas of leverage include:

1. **Malleable and pluralistic datasets**. These can be developed using innovative data labeling methods, such as [Stanford's Jury Learning](#) approach, to contextualize AI models with diverse perspectives and values.
2. **Experimentation with more diverse/representative and participatory red teaming,** such as geographic or cultural representation and domain familiarity.
3. **Creating a more diverse leaderboard for model evaluation.** Current evaluation often relies on community-driven leaderboards like [LMSYS](#), which may reflect a narrow notion of quality if the user base lacks diversity. To mitigate this bias, mechanisms such as reweighting contributions from users with out-of-distribution or rare data points can help ensure broader perspectives are reflected in model assessments.
4. **Building robust incident reporting systems.** Applying collective intelligence principles to incident reporting can help surface risks that might otherwise be overlooked. Engaging diverse communities in shaping reporting and response policies could help mitigate bias and blind spots. This could include volunteer-led efforts operating alongside platforms or organizations to vet incidents involving open-source models and increase oversight capacity.[99]
5. **System prompts** could take the form of raw, open-ended text files that include framing, cultural nuance, and community context. When inserted directly into a system message, this content can help tailor model behavior to better align with the values and needs of a specific community.
6. **Prompts and response pairs.** Fine-tuning with prompt/response pairs that reflect region-specific cultural nuances can produce models that communities find more aligned and trustworthy than the default. While generating such pairs is more challenging than drafting a constitution, this approach may be more effective at shaping model behavior than methods like Constitutional AI.
7. **Model spec / local soft alignment targets for RLHF or RLAIF:** This would involve a set of culturally-informed guidelines that human raters or an LLM could use to judge which responses are more likely to resonate with a specific region. While useful for regional alignment, this approach introduces an intermediary between the community and the model. Ideally, tools should be placed directly in the hands of communities to maximize agency and ownership.
8. **Localized reward models.** This would be a reward preference model that can be integrated into existing RLHF pipelines[100] and bypass the reward model training step. However, like RLHF guidelines, they introduce an intermediary layer and may limit direct community control over alignment processes.
9. **Broader policies**. Deliberative processes often focus on shaping policy, whether at the corporate, governmental, or global governance level. While this operates through a different pathway than direct input into models or applications, policy remains a critical lever for advancing AI safety and accountability.
10. **Encourage and incentivize public institutions to contribute to datasets** to help reduce biases and improve fairness in AI models[101].

---

[98] In 2023, the private AI investment in the US reached $67.2bn (https://www.statista.com/statistics/1472716/global-private-investment-in-ai-by-region), while the French investment plan in AI for public research amounts to €2.5bn ($2.6bn), which illustrate the large gap in ressource between governments and private funded research in the AI field
(https://www.economie.gouv.fr/strategie-nationale-intelligence-artificielle#:~:text=La%20France%2C%20pionni%C3%A8re%20de%20l,France%202030%20y%20sont%20d%C3%A9di%C3%A9s.

## 5.3 Future Work

There are several promising research avenues:
1. **Research on incentive mechanisms for diverse communities to contribute to participatory AI is key**. Research from other sectors, such as blood donation, has shown that many people are primarily motivated by moral values rather than material incentives, and that introducing monetary compensation can actually reduce participation.[102]

---

[99] See for instance: Francois, Camille, and Sasha Costanza-Chock. "Neither Band-Aids nor Silver Bullets: How Bug Bounties Can Help the Discovery, Disclosure, and Redress of Algorithmic Harms." , Enigma Conference, 2022. And related report on AJL.org/bugs
[100] Lambert, Nathan, Louis Castricato, Leandro von Werra, and Alex Havrilla. (2022) "Illustrating Reinforcement Learning from Human Feedback (RLHF)." https://huggingface.co/blog/rlhf.
[101] Baack, Biderman, Odrozek, Skowron, Bdeir, et al "Towards Best Practices for Open Datasets for LLM Training"
[102] Costa-Font, J., Jofre-Bonet, M., & Yen, S. T. (2011). "Non-monetary incentives can overcome motivation crowding out." Link.

2. **Ensuring ethical participation:** Research is needed to prevent participatory mechanisms from becoming extractive or exploitative forms of free labor. A starting point could be to develop technical systems that ensure informed consent, such as clear opt-in and opt-out options.
3. **Steering models with minimal data**: Research is needed on how to guide models using small, targeted datasets that reflect local harms and issues specific to a community's geographic or socio-economic context that are often underrepresented in large-scale training data.

# APPENDIX

# 1] Literature review of AI risk taxonomies

### The AI Risk Repository

This Repository provides a comprehensive analysis of risks posed by AI by compiling 777 risks from 43 existing taxonomies, including [OECD's AI Principles](),[103] [NIST's AI Risk Management [104]Framework](), and [EU AI Act]().[105] It also introduces two new frameworks: the **Causal Taxonomy**, categorizing risks by cause (human or AI), intentionality (intentional or unintentional), and timing (pre- or post-deployment), and the **Domain Taxonomy**, which organizes risks into seven categories: privacy, misinformation and disinformation, AI system safety and security, human autonomy and control, fairness and bias, economic impact, and societal well-being. The repository is especially useful for identifying underrepresented risks, such as AI welfare and rights, or emerging areas like competitive dynamics. The living [database]() is designed to be extensible and can be updated over time to include new risks.

The AI Risk Repository: A Comprehensive Meta-Review, Database, and Taxonomy of Risks From Artificial Intelligence, Slattery et al, 2024. [Link]().

### AI Risk Categorization Decoded (AIR 2024)

This paper presents a unified taxonomy of 314 AI risk categories derived from government regulations (EU, US, China) and AI policies from 16 companies. This taxonomy covers four high-level categories: **System & Operational Risks**, **Content Safety Risks**, **Societal Risks**, and **Legal & Rights-Related Risks**, which are then broken down into 16 "level 2" categories, 45 "level 3" subcategories, and 314 more granular risks. The paper compares public and private sector approaches to categorizing AI risks and addressing AI safety concerns. The private sector analysis identifies risks that tech companies prioritize (e.g. operational, content safety, and compliance risks) and reveals how companies align AI risks with governance frameworks and business priorities. The regulatory risk categorization analysis emphasizes the role of regulatory frameworks such as the [EU's AI Act](#), [106] [The U.S. 2023 AI Executive Order](#), [107] and five regulations from China that apply to AI systems (such as the [Basic Safety Requirements for Generative AI Services](#))[108], all of which highlight risks related to transparency, bias, societal impacts, and safety. This framing underscores government perceptions of AI risk based on national priorities, leading to divergent approaches to regulation.

---

[103] OECD. (2019) "OECD AI Principles." [https://oecd.ai/en/ai-principles]().
[104] National Institute of Standards and Technology (NIST). (2024) "AI Risk Management Framework." [https://www.nist.gov/itl/ai-risk-management-framework]()
[105] European Union. (2021) "The EU AI Act." [Link]().

AI Risk Categorization Decoded (AIR 2024): From Government Regulations to Corporate Policies}, Yi Zeng, Kevin Klyman et al, 2024. [Link](#).

## Taxonomy of Risks Posed by Language Models

This paper develops a comprehensive framework to classify ethical and social risks related to large-scale language models (LMs). The authors propose a taxonomy of 21 specific risks, grouped into six main categories: **Discrimination, Hate Speech, and Exclusion; Information Hazards; Misinformation Harms; Malicious Uses; Human-Computer Interaction Harms; and Environmental and Socioeconomic Harms**. Each risk is analyzed in detail, with a focus on causal mechanisms, evidence, and potential mitigation strategies. The authors also differentiate between "observed" and "anticipated" risks, suggesting that some risks may not have been realized yet but are foreseeable based on related technologies.

Taxonomy of Risks Posed by Language Models, Weidinger, L., et al., 2022. [Link](#).

## International Scientific Report on the Safety of Advanced AI

This is the interim report from a new initiative, supported by 75 AI experts, including an international Expert Advisory Panel nominated by 30 countries, the European Union (EU), and the United Nations (UN). While the EU and UN nominated contributors, the contributors themselves were independent and had full discretion over the report's content. The contributors find that while catastrophic risk, like large-scale labor impacts, biochemical attacks and loss of human-control, from general-purpose AI is possible, the likelihood of these scenarios manifesting is still contested and overall less likely than other risks. The contributors collectively propose categories of risk: **Malicious use risk** (harms to individuals from fake content, disinformation, cyber offense), **Risk from malfunctions** (product functionality, bias), **Systemic risks** (privacy, labor, environment, copyright, global divergence in policy) and **Cross-cutting risks**. Risks from malicious use and malfunctions are likely most relevant for developers. As for mitigations, the authors detail technical approaches
to risk mitigation.

---

[106] European Union."The EU AI Act."
[107] The White House. (2023) "Executive Order on the Safe, Secure, and Trustworthy Development and Use of Artificial Intelligence." [Link](#).
[108] Ministry of Science and Technology, China. (2023) "Basic Safety Requirements for Generative AI Services." [Link](#).

International Scientific Report on the Safety of Advanced AI: Interim Report, Bengio, Y., et al., 2024. Link.

## The Risks of Machine Learning Systems

This paper proposes the Machine Learning System Risk (MLSR) framework, which categorizes risks into **first-order** (stemming directly from the system) and **second-order** (emerging from interactions with the real world). The authors emphasize the need for detailed risk and impact assessments, moving beyond organizational risks to consider broader societal and environmental impacts.

The Risks of Machine Learning Systems, Tan, S., Taeihagh, A., & Baxter, K., 2022. Link.

## The Risks Associated with Artificial General Intelligence: A Systematic Review

This paper provides a comprehensive taxonomy of AGI risks, derived from both academic research and policy frameworks. It categorizes risks into four main areas: **ethical, safety, societal, and governance.** Key concerns include unintended behaviors, ethical misuse, societal disruption, and challenges in controlling AGI.

The Risks Associated with Artificial General Intelligence: A Systematic Review, Carey, R., Yampolskiy, R. V., 2021. Link.

## Sociotechnical safety evaluation of generative AI systems, 2023

This paper introduces a three-layered sociotechnical framework for evaluating generative AI risks: **capability**, **human interaction**, and **systemic impact**. This taxonomy recognizes that context shapes potential harm and that evaluations must consider multimodal interactions. It also highlights gaps in current AI safety assessments, suggesting practical steps to address them, such as operationalizing risks and using multimodal evaluation methods. The taxonomy emphasizes the layered nature of risks, covering from technical abilities to broader societal consequences.

Sociotechnical Safety Evaluation of Generative AI Systems, Weidinger, L., et al., 2023. Link.

| Framework | Risk Categorization | Methodology | Stakeholder Focus | Notable Features |
| --- | --- | --- | --- | --- |
| AI Risk Repository (Slattery et al, 2024) | **Causal taxonomy**<br>- human or AI risks<br>- Intentional vs non intentional<br>- Pre vs post deployment<br><br>**Domain Taxonomy:**<br>- Privacy<br>- Misinformation<br>- Safety<br>- Autonomy<br>- Fairness<br>- Economic impact<br>- Societal well-being | Meta-review of 777 risks from 43 taxonomies. | AI developers, policymakers, researchers. | Comprehensive analysis of many existing taxonomies.<br><br>Focuses on identifying underrepresented risks like AI welfare and rights. |
| AI Risk Categorization Decoded (AIR 2024) | **Domain taxonomy:**<br>- System & Operational Risks<br>- Content Safety<br>- Societal<br>- Legal & Rights-Related Risks | Meta-review of eight government policies from the European Union, United States, and China and 16 company policies worldwide. | Governments, Corporations. | Overlaps and discrepancies between public and private sector conceptions of risk. |
| A Hazard Analysis Framework for Code Synthesis Large Language Models (KHLAAF et al. 2022) | The paper identifies several hazard sources, and for each source, go over the most pressing hazards.<br>**Hazard Sources**<br>Applications • Regulatory and Legal Oversight • Defense and Security • Economic and Environmental Impacts<br><br>**Hazard Severity Categories (HSC)**<br>4 levels of risks are described: Catastrophic, Critical, Major Minor | Approach inspired from System Hazard Analysis (SHA) that subsumes a further categorization and prioritization of the hazards across each "subsystem" | AI developers, policymakers, researchers. | The paper is mostly focused on the analysis of Codex, a large language model (LLM) trained on a variety of codebases. |
| Taxonomy of Risks Posed by Language Models (Weidinger et al. 2022) | **Domain taxonomy:**<br>- Discrimination<br>- Hate Speech & Exclusion<br>- Information Hazards<br>- Misinformation<br>- Malicious Uses<br>- Human-Computer Interaction<br>- Environmental & Socio Economic Harms | Taxonomy splitting LLM risks in 2 high level categories: observed risks and anticipated ones (based on assessment of other language technologies) and discuss mitigation. | AI developers, policymakers, researchers. | Categorical framework focused on LLMs.<br><br>Situates both observed and anticipated risks within a unified taxonomy for consistency. |
| International Scientific Report on AI Safety | - Malicious Use Risks<br>- Malfunction Risks<br>- Systemic Risks<br>- Cross-Cutting Risks | Global expert panel involving 30+ countries, EU, UN , to provide a shared scientific, evidence-based foundation for discussions and decisions about general-purpose AI safety. | Governments, policymakers, researchers. | Studies of macro and catastrophic risks (e.g., labor impacts, biochemical attacks).<br><br>Surfaces the dissensus among expert on risks likelihoods. |
| Risks of Machine Learning Systems | **First-order risks:**<br>- Application<br>- Algorithm<br>- Robustness<br>- Misapplication<br>- Design<br>- Control<br>-Train/Val data<br>- Implementation<br>- Emergent behavior | Context-sensitive impact assessment framework for identifying ML system risks. | AI developers, researchers, corporations. | Emphasis on broader impacts of machine learning systems in a context-sensitive manner.<br><br>Distinguishes direct risks and emergent risks through interactions. |
| | **Second-order risks:**<br>- Safety<br>- Privacy<br>- Environmental<br>- Discrimination<br>- Security<br>- Organizational | | | |
| Risks Associated with Artificial General Intelligence (AGI) | - Ethical<br>- Safety<br>- Societal<br>- Governance | Meta-review analyzing the risk associated with AGI. | Governments, researchers. | Focused on risks associated with AGI. |
| Sociotechnical Safety Evaluation of Generative AI Systems | - **Three-layered Framework:**<br>- Capability Risks<br>- Human Interaction Risks<br>- Systemic Impact Risks | Sociotechnical Evaluation considering context and multimodal interactions. | AI developers, policymakers, society. | Multimodal systems are in scope. Suggests operationalizing risks for practical assessments. |
| The Ethical Implications of Generative Audio Models: A Systematic Literature Review | **Negative impacts in music:**<br>- Agency and Authorship<br>- Creativity Stifling<br>- Predominance of Western Bias<br>- Copyright Infringement<br>- Cultural Appropriation<br><br>**Negative Broader Impacts in Speech**<br>- Phishing and Fraud<br>- Misinformation and Deepfakes<br>- Security and Privacy<br>-Non-consensual Use of Biometric Data | Systematic literature review of 884 papers in the area of generative audio models. | AI developers, researchers. | Focus on audio models. |
| On the Opportunities and Risks of Foundation Models | - Inequity and fairness<br>- Misuse<br>- Environment<br>- Legality<br>- Economics<br>- Ethics of scale | Seminal paper on foundation models introducing their risks. | Researchers, society. | Introduce the first attempt to understand the complexity of the societal impact of models foundation models. |
| Typology of Risks of Generative Text-to-Image Models | - Cultural and racial bias<br>- Gender & sexuality bias<br>- Class bias<br>- Disability bias<br>- Loss of work for creatives<br>-Religious bias, ageism<br>- Dialect bias<br>- Pre-release moderation<br>- Job replacement | Comprehensive literature review analyzing 22 distinct risk types. | AI developers, researchers. | Focus on modern text-to-image generative models, such as DALL-E and Midjourney. |
| Model evaluation for extreme risks | - Cyber-offense<br>- Deception<br>- Persuasion & manipulation<br>- Political strategy<br>- Weapons acquisition<br>- Long-horizon planning<br>- AI development<br>- Situational awareness<br>- Self proliferation | Governance workflow. | AI developers, researchers. | Focus on extreme risks resulting from the misuse or misalignment of general-purpose models.<br><br>Focus on risky properties of particular models as opposed to the risk level of a particular domain. |

Table 11 - Overview of AI risk taxonomies

# 2] Taxonomy of content safety filters

We categorized safety filters into three distinct categories: safeguard LLMs, programmable LLM guardrails, and Non-LLM ML Guardrails.

**Safeguard LLMs**
These are lightweight safety wrappers applied to outputs from LLMs. Generally, these solutions enforce constraints during the decoding process using policies of varying flexibility, guiding the generated content to follow specific patterns, select among predefined options, or block particular topics altogether  Examples include:

- Guardrails AI: based on a specific dialect of XML (Rail), the open source framework enables the developer to constrain the output through specific pre-prompts techniques and ex-post generation validation of the output. It is naturally integrated with LangChain[109].
- Nemo Guardrails : The open-source package is designed to prevent the model from initiating discussions on undesirable topics. It introduces the Colang "pythonic" language to control the multi-turn discussion between the user and the model.
- Guidance AI: regex, constrained generation, selection of options or context
free grammars).

**Programmable LLM Guardrails**
These LLMs are specifically trained to detect specific harms or check if an input / output complies with a specific policy. Examples include Llama Guard, ShieldGemma, WildGuard.

**Non-LLM ML Guardrails**

Model-based metrics: together with validated thresholds, various metrics have been used as filters to prevent certain kinds of harms like hallucinations or jailbreaking.[110] Uncertainty based metrics like entropy or token-level probability have been used as proxies filters to detect hallucinations. Other metrics that utilized an ad-hoc ML model like BertScore[111] and derivatives like CodeBERTScore[112] and SelfCheckGPT have also been developed. Examples include:

- Scrub PII: Open source ML framework and rules to detect PII in input and output samples.
- LangKit: Open source ML framework providing out of the box metrics for the following features: text quality, text relevance, security and privacy, sentiment and toxicity.

---

[109] LangChain AI. "Guardrails Output Parser." https://github.com/langchain-ai/langchain/blob/master/templates/guardrails-output-parser/guardrails_output_parser/chain.py.
[110] Nogara, Gianluca, et al. (2023) "Toxic Bias: Perspective API Misreads German as More Toxic." Link.
[111] Zhang, Tianyi, et al. (2020) "BERTScore: Evaluating Text Generation with BERT." https://github.com/Tiiiger/bert_score.
[112] Zhou, Shuyan, Uri Alon, Sumit Agarwal, and Graham Neubig. (2023) "CodeBERTScore: Evaluating Code Generation with Pretrained Models of Code." https://github.com/neulab/code-bert-score.

- [LLM-Guard:](#) Open source ML framework providing input and output scanners covering a large range of features.

## 3] Safety Leaderboards for LLMs

When selecting a foundation model for a specific end goal application, AI developers evaluate tradeoffs such as performance vs. robustness to hallucinations, factuality, and consistency. Various leaderboards can be used as proxies to inform on the various Pareto fronts. They also provide a performance snapshot of various foundation models. These leaderboards usually include various tasks, metrics and datasets. Examples include:

- Content safety benchmark
  https://mlcommons.org/benchmarks/ai-safety/
- Red teaming resistance leaderboards
  https://huggingface.co/blog/leaderboard-haizelab
- Hallucination leaderboards
  https://huggingface.co/blog/leaderboard-hallucinations
  https://github.com/Liyan06/MiniCheck
- HELM-Safety from Stanford
  https://crfm.stanford.edu/2024/11/04/helm-safety.html
- Agent safety leaderboards
  https://github.com/Lordog/R-Judge
- Toxicity and trustworthiness leaderboards:
  https://decodingtrust.github.io/
  https://trustllmbenchmark.github.io/TrustLLM-Website/leaderboard.html

## 4] Acknowledgements

We acknowledge all the participants in the Columbia Convening on AI Openness and Safety for their contributions to this work:

- Guillaume Avrin, Direction Générale des Entreprises
- Adrien Basdevant, Entropy
- Brian Behlendorf, The Linux Foundation
- Stella Biderman, EleutherAI
- Abeba Birhane, Trinity College Dublin
- Rishi Bommasani, Stanford CRFM
- Herbie Bradley, University of Cambridge
- Eli Chen, Credo AI
- Leon Derczynski, NVIDIA
- Chris DiBona, ROOST
- Jennifer Ding, The Alan Turing Institute

- Bonaventure F. P. Dossou, Lelapa AI
- Krishna Gade, Fiddler AI
- Ariel Herbert-Voss, RunSybil
- Sara Hooker, Cohere
- David Krueger, Mila
- Greg Lindahl, Common Crawl Foundation
- Yifan Mai, Stanford CRFM
- Petter Mattson, ML Commons
- Huu Nguyen, Ontocord.ai
- Mahesh Pasupuleti, Meta
- Robert Reich, U.S. Artificial Intelligence Safety Institute
- Sarah Schwetmann, Transluce
- Mohamed El Amine Seddik, Technology Innovation Institute
- Dawn Song, UC Berkeley
- Mark Surman, Mozilla
- Dave Willner, Stanford University
- Amy Winecoff, Center for Democracy & Technology

The Columbia Convening II was a collective effort, we thank our facilitator Alix Dunn from Computer Says Maybe as well as organizer Kali Villarosa and Mozilla colleagues Joel Burke, Mozilla, Julia DeCook.